\documentclass[pdflatex,sn-mathphys]{sn-jnl}% Math and Physical Sciences Reference Style
\usepackage{makecell}
\usepackage[T1]{fontenc}
\jyear{2023}%

\theoremstyle{thmstyleone}%
\theoremstyle{thmstyletwo}%

\theoremstyle{thmstylethree}%

\begin{document}

\title[Quantifying Impairment and Disease Severity Using AI]{Quantifying Impairment and Disease Severity Using AI Models Trained on Healthy Subjects}


\author[1]{\fnm{Boyang} \sur{Yu}}\email{boy.yu@nyu.edu}

\author[1]{\fnm{Aakash} \sur{Kaku}}\email{ark576@nyu.edu}

\author[1]{\fnm{Kangning} \sur{Liu}}\email{kl3141@nyu.edu}

\author[3,4]{\fnm{Avinash} \sur{Parnandi}}\email{avinashparnandi@gmail.com}

\author[3]{\fnm{Emily} \sur{Fokas}}\email{emily.fokas@nyulangone.org}

\author[3]{\fnm{Anita} \sur{Venkatesan}}\email{anita.venkatesan@nyulangone.org}

\author[3]{\fnm{Natasha} \sur{Pandit}}\email{ngp238@nyu.edu}

\author[1,2]{\fnm{Rajesh} \sur{Ranganath}}\email{rajeshr@cims.nyu.edu}
\equalcont{These authors contributed equally to this work.}
\author*[3,4]{\fnm{Heidi} \sur{Schambra}}\email{Heidi.Schambra@nyulangone.org}
\equalcont{These authors contributed equally to this work.}
\author*[1,2]{\fnm{Carlos} \sur{Fernandez-Granda}}\email{cfgranda@cims.nyu.edu}
\equalcont{These authors contributed equally to this work.}

\affil[1]{\orgdiv{Center for Data Science}, \orgname{New York University, NYU}, \orgaddress{\street{60 Fifth Ave}, \city{New York}, \postcode{10011}, \state{New York}, \country{United States}}}

\affil[2]{\orgdiv{Courant Institute of Mathematical Sciences}, \orgname{NYU}, \orgaddress{\street{ 251 Mercer St}, \city{New York}, \postcode{10012}, \state{New York}, \country{United States}}}

\affil[3]{\orgdiv{Department of Neurology}, \orgname{NYU Grossman School of Medicine}, \orgaddress{\street{ 550 1st Ave}, \city{New York}, \postcode{10016}, \state{New York}, \country{United States}}}

\affil[4]{\orgdiv{Department of Rehabilitation Medicine}, \orgname{NYU Grossman School of Medicine}, \orgaddress{\street{ 550 1st Ave}, \city{New York}, \postcode{10016}, \state{New York}, \country{United States}}}


\abstract{Automatic assessment of impairment and disease severity is a key challenge in data-driven medicine. We propose a novel framework to address this challenge, which leverages AI models trained exclusively on healthy individuals. %The models are designed to predict a clinically-meaningful attribute of the healthy patients. When presented with data where the attribute is affected by the medical condition of interest, the models experience a decrease in confidence. 
The COnfidence-Based chaRacterization of Anomalies (COBRA) score exploits the decrease in confidence of these models when presented with impaired or diseased patients to quantify their deviation from the healthy population. We applied the COBRA score to address a key limitation of current clinical evaluation of upper-body impairment in stroke patients. The gold-standard Fugl-Meyer Assessment (FMA) requires in-person administration by a trained assessor for 30-45 minutes, which restricts monitoring frequency and precludes physicians from adapting rehabilitation protocols to the progress of each patient. The COBRA score, computed automatically in under one minute, is shown to be strongly correlated with the FMA on an independent test cohort for two different data modalities: wearable sensors ($\rho = 0.845$, 95\% CI [0.743,0.908]) and video ($\rho = 0.746$, 95\% C.I [0.594, 0.847]). To demonstrate the generalizability of the approach to other conditions, the COBRA score was also applied to quantify severity of knee osteoarthritis from magnetic-resonance imaging scans, again achieving significant correlation with an independent clinical assessment ($\rho = 0.644$, 95\% C.I [0.585,0.696]).  
}




\maketitle

\section{Introduction}\label{sec:intro}

%Understanding the differences between patients and healthy individuals is crucial. With precise and objective illness measurement, healthcare providers can gain insights into underlying causes and develop more effective treatment plans.
% The assessment of impairment and disease severity is a key challenge in medicine. 

In current clinical practice, assessment of impairment and disease severity typically relies on examinations by medical professionals~\cite{medsger2003assessment,fugl1975method}. As a result, assessment is often qualitative and its frequency is constrained by clinician availability. Developing data-driven quantitative metrics of impairment and disease severity has the potential to enable continuous and objective monitoring of patient recovery or decline. Such monitoring would facilitate personalized treatment and administration of appropriate therapeutic interventions in telehealth and other remotely supervised contexts where ongoing access to clinicians is not readily available~\cite{raman2023machine,hwangbo2022machine,shamout2021artificial}.

%, which are may be too costly and time consuming and often qualitative rather than quantitative. Quantitative data-driven measures of impairment and disease severity have the potential to facilitate the design of personalized treatment plans, systematic tracking of patient progress, and provide appropriate therapeutic interventions. 

Artificial-intelligence (AI) models based on machine learning are a natural tool to perform data-driven patient assessment~\cite{cottrell2017real, laver2020telerehabilitation,hamet2017artificial, palanica2020need,ting2020artificial, topol2019high, barnes2021artificial,jeddi2020remote,shaik2023remote,sawyer2020wearable,
akbilgic2019machine,chen2023prediction,babenko2022detection,shen2021interpretable}. These models can be trained in a supervised fashion to estimate labels associated with patient data from large curated datasets of examples~\cite{shaik2023remote,topol2019high, beam2018big}. Unfortunately, it is often very challenging to assemble datasets containing an exhaustive representation of severity or impairment levels, which is necessary to ensure the accuracy of the AI models~\cite{ching2018opportunities,norori2021addressing, van2001functional,langs2013visceral,oakden2020hidden,roy2022demystifying}. Moreover, supervised approaches require the existence of an objective quantitative metric that can be computed for every patient in the dataset, but such metrics do not exist for many medical conditions~\cite{jarrett2019applications,varoquaux2022machine}. 
 
To address these challenges, we consider the problem of performing automatic patient assessment using AI models trained 
% rr: is healthy a binarization of the patient assessment?
\emph{only on data from healthy subjects}. This is an anomaly detection problem, where the goal is to identify data points that are systematically different from a reference population~\cite{chandola2009anomaly}. Existing anomaly-detection methods for medical data are mostly based on generative models~\cite{akcay2019ganomaly,deecke2019image}. These models are designed to reconstruct high-dimensional data from a learned low-dimensional representation. Once trained, they are typically unable to accurately reconstruct data that are anomalous, due to their inconsistency with the training set. Consequently, the model reconstruction error tends to be higher for anomalies than for normal data, and can therefore be used as an anomaly-detection score. This approach has been applied to identify chronic brain infarcts~\cite{van2021anomaly}, Alzheimer's disease~\cite{pinaya2021using}, microstructural abnormalities in diffusion MRI tractometry~\cite{chamberland2021detecting}, and abnormalities of cosmetic breast reconstruction in cancer patients~\cite{kim2022feasibility}. 

\begin{figure}[t!]%
\centering
\includegraphics[width=0.9\textwidth]{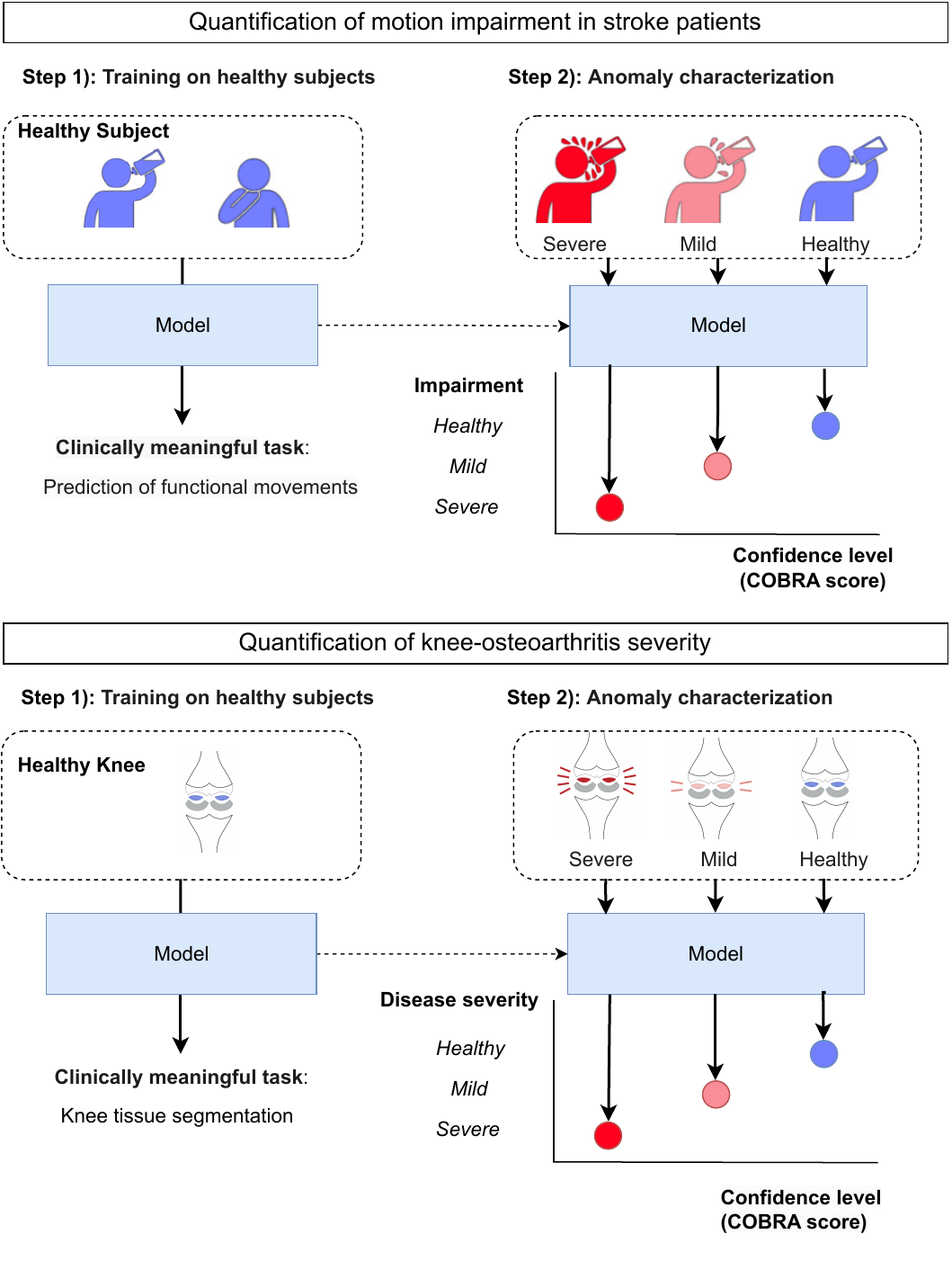}
\caption{\textbf{The COnfidence-Based chaRacterization of Anomalies (COBRA) score.} In Step 1, an AI model is trained to perform a clinically meaningful task on data from healthy individuals. For impairment quantification in stroke patients, the task is prediction of functional primitive motions from videos or wearable-sensor data (top). For severity quantification of knee osteoarthritis, the task is segmentation of knee tissues from magnetic resonance imaging scans (bottom). In Step 2, the COBRA score is computed based on the confidence of the AI model when performing the task on patient data. Data from patients with higher degrees of impairment or severity differ more from the healthy population used for training, which results in decreased model confidence and hence a lower COBRA score.
%are trained to recognize how healthy population behave in clinically meaningful tasks, such as movement patterns during drinking water. In order to characterize anomaly, we quantify subjects' deviations from the healthy reference by calculating the drop in model confidence. Higher the severity impairment level, larger the drop in confidence level. 
}\label{fig:COBRA}
\end{figure}

Anomaly detection based on generative models has an important disadvantage: it does not constrain the AI model to learn clinically relevant features. 
%First, it assumes the existence of an AI model capable of reconstructing the data, which may not be available for certain data modalities, such as wearable-sensor data. Second, the AI model is not constrained to learn clinically-relevant features. 
Consequently, the model reconstruction error may depend on properties of the data unrelated to the medical condition of interest. Here, we propose a novel anomaly-detection framework that is \emph{tailored to a specific medical condition}. This is achieved by utilizing an AI model that predicts an attribute of the data, which is directly relevant to the condition (e.g. type of motion primitive performed by the stroke-impaired side, or tissue type for knee osteoarthritis). Crucially, the model is trained exclusively on healthy subjects, using annotated data describing the attribute. 
% rr: is the type of motion primitive affected bt the condition
When the models are presented with data where the attribute is affected by the condition, we observe that the average model confidence tends to decrease proportionally to severity. %This loss of confidence can be used to detect such individuals and quantify to what extent they differ from the healthy population. 
This yields a quantitative patient-assessment metric, which we call the COnfidence-Based chaRacterization of Anomalies (COBRA) score. %The approach is analogous to how clinicians build an internal reference of what a healthy structure should look like. 
Figure~\ref{fig:COBRA} provides a schematic description of the proposed framework.  

\begin{figure}
\begin{tabular}{>{\centering\arraybackslash}m{0.3\linewidth} > {\centering\arraybackslash}m{0.3\linewidth} > {\centering\arraybackslash}m{0.3\linewidth}}
% \multicolumn{2}{c}{(a)Quantification of motion impairment in stroke patients}  \\
\small{Wearable sensors} & \small{Video} & \small{MRI scan}\\
\includegraphics[width=1.13\linewidth]{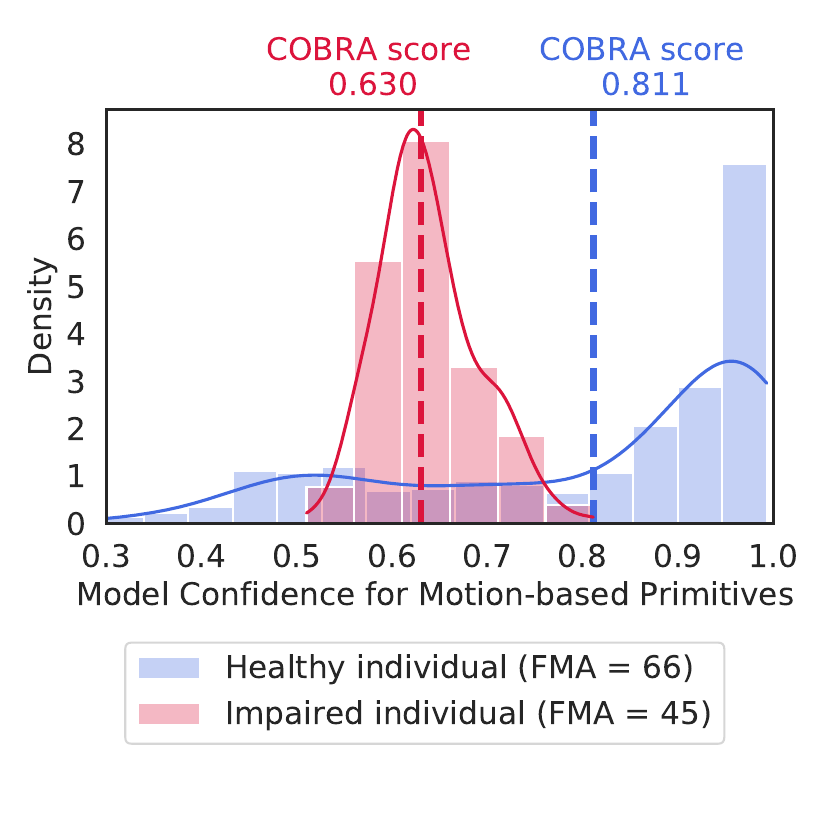} & \includegraphics[width=1.13\linewidth]{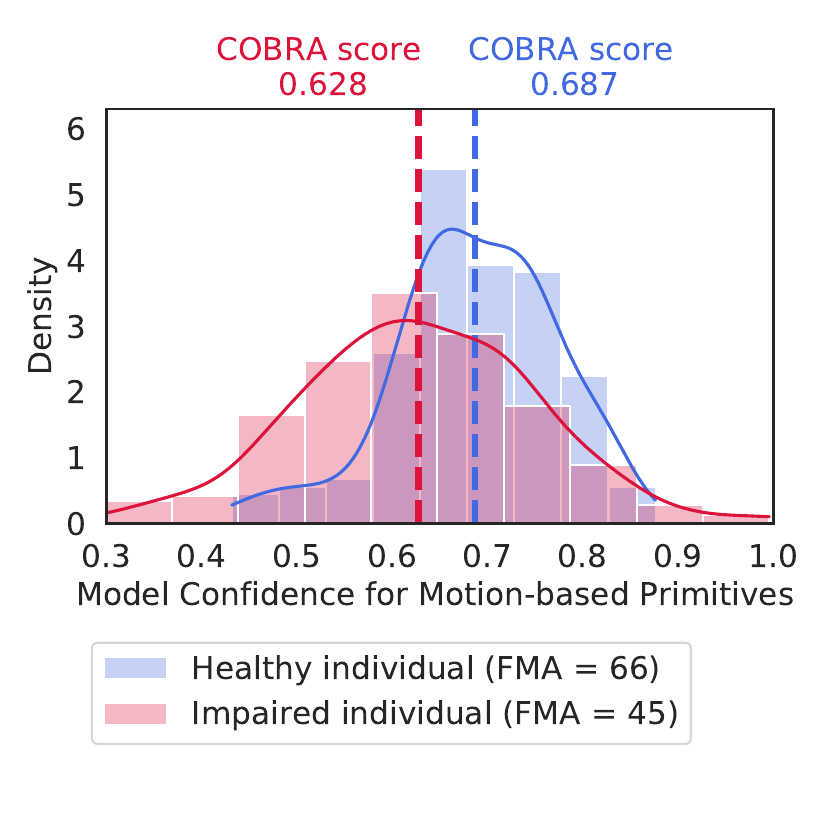} & \includegraphics[width=1.13\linewidth]{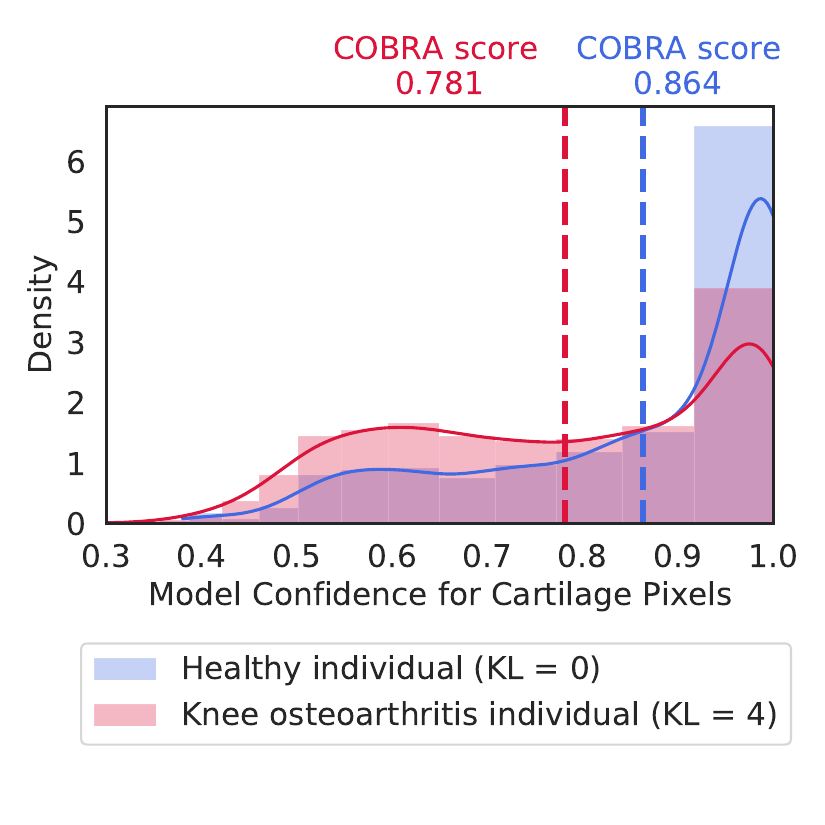}\\
\end{tabular}
\caption{
% rr: this figure doesn't actually show that the average is disciminative because it's only a single sample
\textbf{Averaging model confidence yields a discriminative subject-level metric.} The plots show histograms and kernel density estimates of the confidence of a model trained on healthy subjects when presented with test data from an impaired or diseased patient (red), and from a held-out healthy individual (blue). The confidence distributions overlap, so individual values do not allow to discriminate between healthy and impaired individuals. In contrast, the average confidence is systematically higher for healthy subjects, and therefore provides a discriminative subject-level metric. The first and second plot correspond to wearable-sensor and video data associated with the same healthy and impaired individuals from the test cohort for quantification of stroke-induced impairment. The third plot corresponds to MRI scans from a healthy and diseased individual in the knee-osteoarthritis test cohort.
%wearable-sensor data: Healthy subject id = c30 Stroke subject id= s48, 
%video data: Healthy subject id = c23, Stroke subject id= s30 ,
% knee data: Healthy subject id = c23, Stroke subject id= s30 
}\label{fig:COBRA_histograms}
\end{figure}

The COBRA score is inspired by a technique proposed in~\cite{hendrycks2016baseline}, which identifies anomalous data points using the confidence of AI models. In this and subsequent works \cite{chen2020robust,hsu2020generalized,vyas2018out,mohseni2020self,devries2018learning} anomalies were identified based on the loss of confidence of the AI models for a \emph{single data point}. 
The effectiveness of this approach depends on the overlap in the distribution of confidences \citep{zhang2021understanding}.
In our applications of interest, this is ineffective, as illustrated by Figure~\ref{fig:COBRA_histograms}. When presented with multiple inputs from an impaired or diseased patient, AI models trained on healthy subjects tend to lose confidence \emph{on average}, but the confidence for individual data points is very noisy and results in an unreliable metric. For this reason, the COBRA score is computed using multiple data points for each subject, corresponding to different motions in the application to stroke and to different pixels in the application to knee osteoarthritis. Aggregating the confidence associated with multiple data via averaging dramatically reduces the noise, resulting in a stable and accurate subject-level metric.

We apply the COBRA score to automatically evaluate the impairment level of stroke patients. Stroke commonly causes motor impairment in the upper extremity (UE), characterized by loss of strength, precise control, and intrusive
muscle co-activation, which collectively interfere with normal function. Rehabilitation seeks to reduce motor impairment through the repeated practice of functional movements with the UE. In this process, it is crucial to monitor the impairment level of the patient. The gold-standard method of measuring motor impairment is the Fugl-Meyer Assessment (FMA)~\cite{fugl1975method}.  Unfortunately, it requires in-person administration by a trained assessor and is time-consuming (30-45 minutes), which makes it impractical for frequent monitoring. Automatic assessment of motion impairment based on video or wearable-sensor data would address these limitations, facilitating actionable and granular tracking of motor recovery. 

Motor impairment evaluation in stroke patients illustrates the difficulty of applying standard supervised AI methodology to patient assessment. An existing study shows the feasibility of the approach~\cite{park2020automatic}, but only includes 17 patients. Training a supervised model to predict impairment and rigorously evaluating its performance on held-out data requires a database of at least hundreds, and ideally thousands of patients, labeled with the corresponding impairment level. However, the largest such publicly available dataset consists of just 51 patients~\cite{kaku2022strokerehab}. Here, we use this dataset as a held-out test set to evaluate the proposed framework.  

In order to assess impairment in stroke patients using the COBRA score, we trained AI models  to predict classes of UE motion, known as functional primitives, performed by a group of healthy individuals. We used upper-body motion data obtained from videos and wearable sensors. The model was then applied to data from a test cohort of stroke patients and held-out healthy subjects performing nine different stroke rehabilitation activities. The confidence of the motion predictions for each test subject was averaged to compute the corresponding COBRA score. 
As reported in Section~\ref{sec:results_stroke}, for both data modalities the COBRA score is correlated with the Fugl-Meyer Assessment of the patients, obtained in person by trained experts. This greatly expands on our preliminary findings, which used a similar approach with wearable-sensor data from a single rehabilitation activity~\cite{parnandi2023data}.  

To demonstrate the general applicability of the COBRA framework, we show that it can be used to evaluate severity of knee osteoarthritis from magnetic resonance imaging (MRI) scans. Knee osteoarthritis is a musculoskeletal disorder  characterized by a progressive loss of knee cartilage. To quantify severity, we trained an AI model to perform segmentation of different knee tissues (including cartilage) on MRIs of healthy knees. We then applied the model to knee MRIs from a test cohort of diseased patients and held-out healthy subjects. The confidence of the tissue predictions for each test subject was averaged to compute the corresponding COBRA score.  As reported in Section~\ref{sec:results_knee}, the resulting COBRA score is again highly correlated with an independent assessment of disease severity (in this case, the Kellgren-Lawrence grade).

\section{Results}\label{sec:results}

\subsection{Quantification of Impairment in Stroke Patients}
\label{sec:results_stroke}
\begin{table}[t]
\begin{center}
\begin{minipage}{314pt}
\caption{Demographic and clinical characteristics of the training and test cohorts for the application to quantification of motion impairment in stroke patients. The mean $\pm$ standard deviation is reported for age, Fugl-Meyer assessment and time since stroke.}\label{tab:stroke_cohort}%
\resizebox{1\linewidth}{!}{
\begin{tabular}{lll}
\toprule
   & Training  & Testing \\
\midrule
Number of subjects & 25   & 55  \\
Trials    & 1265   & 2183  \\
Age   & 62.4 $\pm$ 13.1    & 57.7 $\pm$ 14.0  \\
Sex  & 13 male, 12 female    & 25 male, 30 female  \\
Race\footnotemark[1]  & 10 W, 12 B, 0 A, 1 AI, 2 O   & 24 W, 14 B, 9 A, 0 AI, 8 O   \\
Paretic Side & n/a &  28 left, 23 right, 4 n/a  \\
Fugl-Meyer Assessment  & 66    & 43.5 $\pm$ 16.2  \\
Impairment level\footnotemark[2]   & 25 healthy  & 4 healthy, 20 mild, 23 moderate, 8 severe   \\
Time since stroke   & n/a  & 5.4 $\pm$ 6.1 (for stroke patients) \\
\botrule
\end{tabular}
}
% \footnotetext{Source: This is an example of table footnote. This is an example of table footnote.}
%\footnotetext[1]{Sex: Male(M), Female(F)}
\footnotetext[1]{Race: White (W), Black (B), Asian (A), American Indian (AI), Other (O)}
%\footnotetext[3]{Paretic Side: Left(L), Right(R)}
%\footnotetext[4]{Fugl-Meyer Assessment(FMA): range 0-66, with 66 indicating healthy}
\footnotetext[2]{Based on FMA: 0-25 is severe, 26-52 moderate, 53-65 mild, and 66 healthy.}
\end{minipage}
\end{center}
\end{table}

\textbf{Data}. The application of the COBRA score to the impairment quantification in stroke patients was carried out using the publicly available StrokeRehab dataset~\cite{kaku2022strokerehab}. StrokeRehab contains video and wearable-sensor data from a cohort of 29 healthy individuals and 51 stroke patients performing multiple trials of nine rehabilitation activities (described in Tables~\ref{tab:act_descp_1} and \ref{tab:act_descp_2}). The  impairment level of each patient was quantified via the Fugl-Meyer assessment (FMA)~\cite{fugl1975method}. The FMA score is a number between 0 (maximum impairment) and 66 (healthy) equal to the sum of itemized scores (each from 0 to 2) for 33 upper body mobility assessments carried out in-clinic by a trained expert. The wearable-sensor and video data are labeled to indicate what functional primitive is being performed by the paretic arm over time: reach (UE motion to make contact with a target object), reposition (UE motion to move into proximity of a target object), transport (UE motion to convey a target object in space), stabilization (minimal UE motion to hold a target object still), and idle (minimal UE motion to stand at the ready near a target object). 
%are described in Section~\ref{sec:methods_stroke_data}. % Additional details about the data are provided in Section~\ref{sec:methods_stroke_data}.

%which our analysis was predicated upon, is a large-scale action-recognition dataset that includes subtle short-duration actions (\emph{i.e.} fundamental functional primitives) labeled at a high temporal resolution. This multimodal dataset incorporates recordings from wearable sensors and cameras, capturing participants undertaking a myriad of daily activities. For each activity, each participant may perform 1 to 10 trials. In each trial, subjects perform a set of fundamental functional primitives such as reaches, transports, repositions, stabilizations, and idles~\cite{schambra2019taxonomy}. The dataset focuses on the upper extremity (UE) and includes UE FMA scores~\cite{fugl1975method} assigned by a trained assessor. The subjects, both healthy and stroke-affected, are stratified based on their FMA scores as follows: healthy (FMA score 66), mild (FMA score 53–65), moderate (FMA score 26–52), severe (FMA score 0–25)~\cite{woodbury2013rasch}. Additional details about the data can be found in Section~\ref{sec:methods_stroke_data}.

\noindent \textbf{Computation of COBRA score}.  AI models were trained to predict the functional primitives performed by a training cohort consisting of 25 of the 29 healthy individuals (selected at random). The model input was either wearable sensor or video data. Detailed descriptions of these models are provided in Section~\ref{sec:COBRA_stroke}. The models were applied to a test cohort consisting of the remaining 4 healthy individuals and the 51 stroke patients. Demographic and clinical information about the training and test cohorts is provided in Table~\ref{tab:stroke_cohort}. The COBRA score was calculated as the average of the model confidence for data points identified by the models as corresponding to functional primitives that involve motion (\emph{transport}, \emph{reposition}, and \emph{reach}), as described in Section~\ref{sec:COBRA_stroke}.

\begin{figure}
\begin{tabular}{
 >{\centering\arraybackslash}m{0.45\linewidth} >{\centering\arraybackslash}m{0.45\linewidth}  }
 \multicolumn{2}{c}{(a) Motion impairment in stroke patients} \vspace{0.2cm}\\
 Wearable Sensors & Video \\
 \makecell{$\rho$ = 0.814\\{\small 95$\%$ CI [0.700,0.888]}}   & \makecell{$\rho$ = 0.736\\{\small 95$\%$ CI [0.584,0.838]}} \\ 
\includegraphics[width=0.85\linewidth]{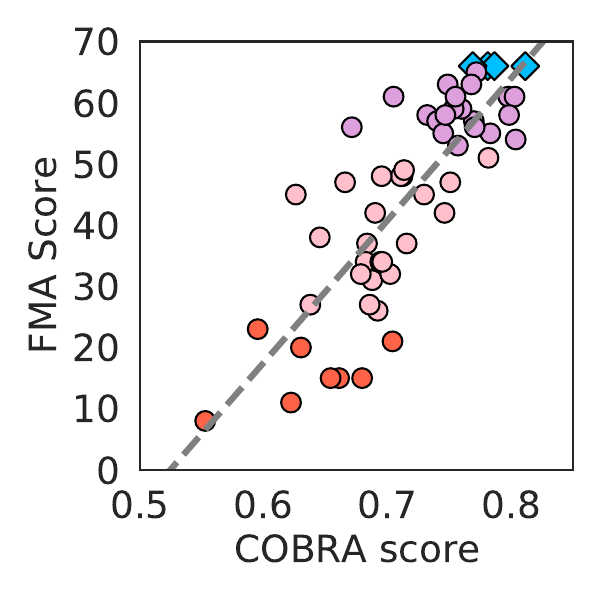} &   \includegraphics[width=0.85\linewidth]{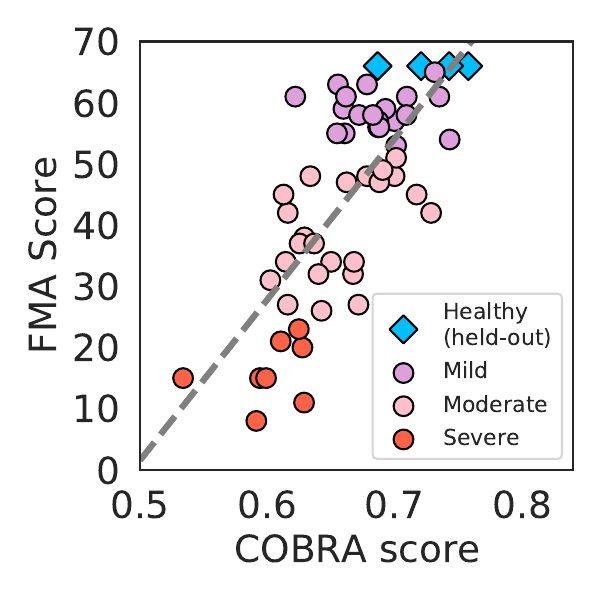} \\
 \multicolumn{2}{c}{(b) Severity of knee-osteoarthritis from MRI scans}\\
 \multicolumn{2}{c}{\makecell{$\rho$ = -0.644\\{\small 95$\%$ CI [-0.696, -0.585]}} }\\
 \multicolumn{2}{c}{\includegraphics[width=0.9\linewidth]{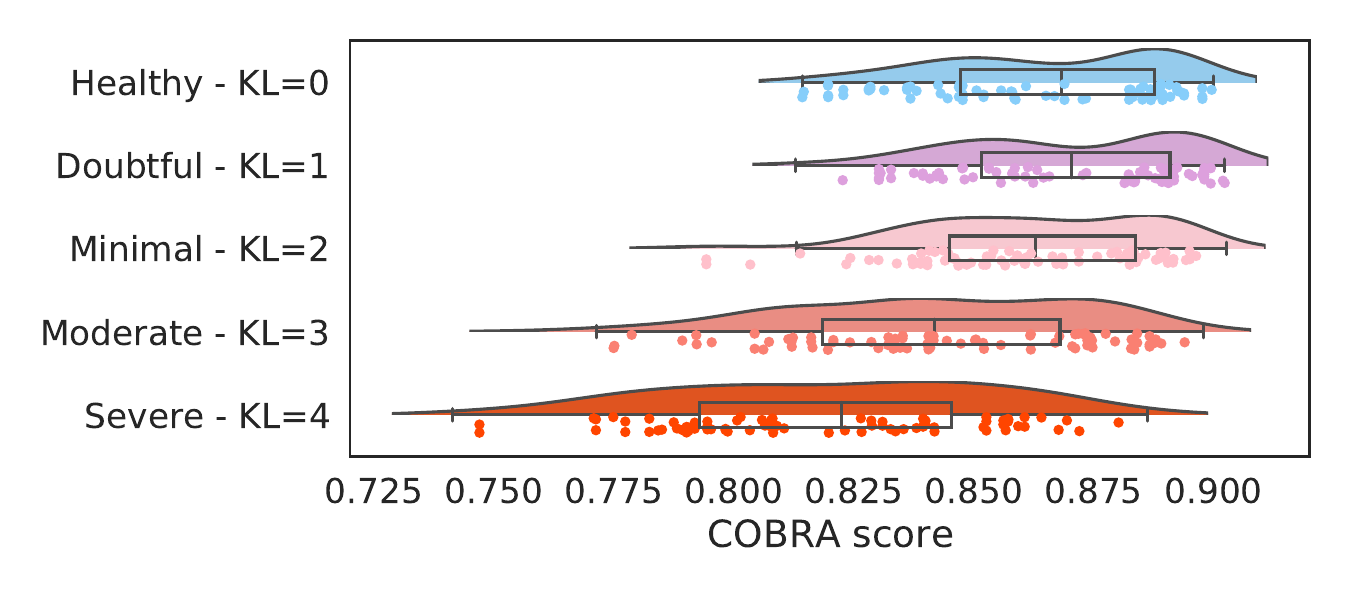}} \\
\end{tabular}
\caption{\textbf{Correlation between the COBRA score and clinical assessment.} The top row shows scatterplots of the Fugl-Meyer assessment (FMA) score, based on in-person examination by an expert, and the proposed data-driven COBRA score computed from wearable-sensor (left) and video (right) data. The correlation coefficient $\rho$ between the two scores is high, particularly for the wearable-sensor data. The second row shows a scatterplot and density plots of COBRA scores computed from magnetic-resonance imaging (MRI) knee scans of patients with different Kellgren-Lawrence (KL) grades. The KL grade and the COBRA score exhibit significant inverse correlation.
%capture motion impairment in stroke patients. Results on wearable sensor (left) and video (middle) data both have strong correlation with patients' FMA scores.
%Adopting single structured activity (shelf or RTT) for calculating COBRA achieves comparable quantification results (right) while using far fewer data.
%(b) Quantification of knee osteoarthritis severity using MRI images using COBRA show consistency with the Kellgren and Lawrence (KL) grading system. Each cluster represents COBRA scores distribution for patients with the same KL values. 
}\label{fig:COBRA_scatterplots}
\end{figure}

\noindent \textbf{Evaluation}. 
The COBRA score was evaluated by computing its Pearson correlation coefficient with the Fugl-Meyer Assessment (FMA) score~\cite{fugl1975method} on the test cohort of 51 stroke patients and 4 healthy individuals (n=55). The correlation coefficient is 0.814 (95\% CI [0.700,0.888]) for the wearable-sensor data and 0.736 (95\% CI [0.584, 0.838]) for the video data. Figure \ref{fig:COBRA_scatterplots} (a) shows  scatterplots of the COBRA and FMA scores. For both data modalities, the COBRA score has a strong, statistically significant correlation with the in-clinic assessment.

%between the final COBRA scores and the Fugl-Meyer Assessment (FMA) scores of stroke patients is depicted in Figure \ref{fig:COBRA_scatterplots}(a). Our analysis reveals a strong correlation between COBRA scores and patients' FMA scores regardless of the input modalities–specifically, a Pearson correlation coefficient of 0.814 (Confidence Interval (CI) [0.700,0.888]) for wearable sensor input, and 0.736 (C.I [0.584, 0.838]) for video input. These findings lend support to our hypothesis that model confidence accurately reflects the extent of impairment. 

\begin{figure}
\begin{tabular}{>{\centering\arraybackslash}m{0.47\linewidth} >{\centering\arraybackslash}m{0.47\linewidth}}
 Wearable Sensors   & Video \\
 \includegraphics[width=1.\linewidth]{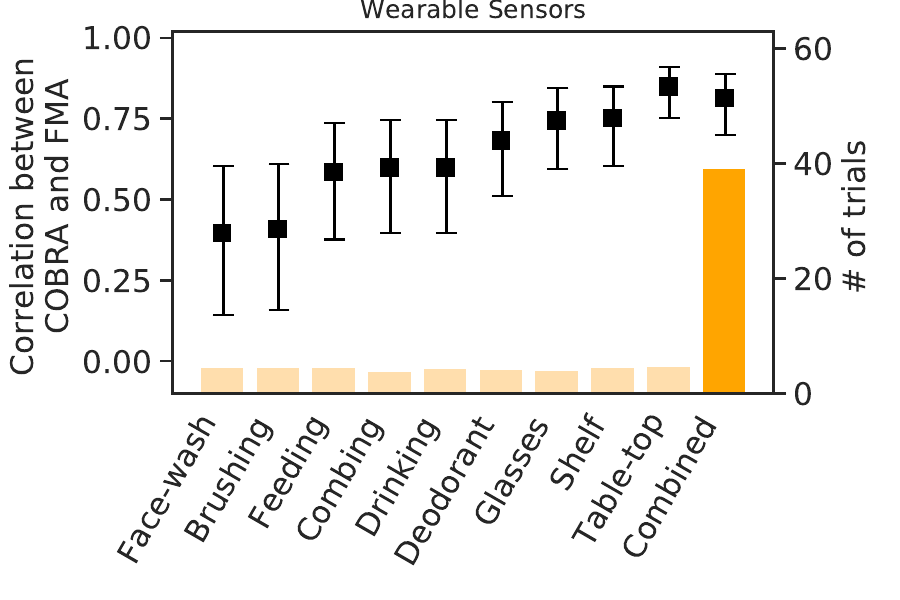} & \includegraphics[width=1.\linewidth]{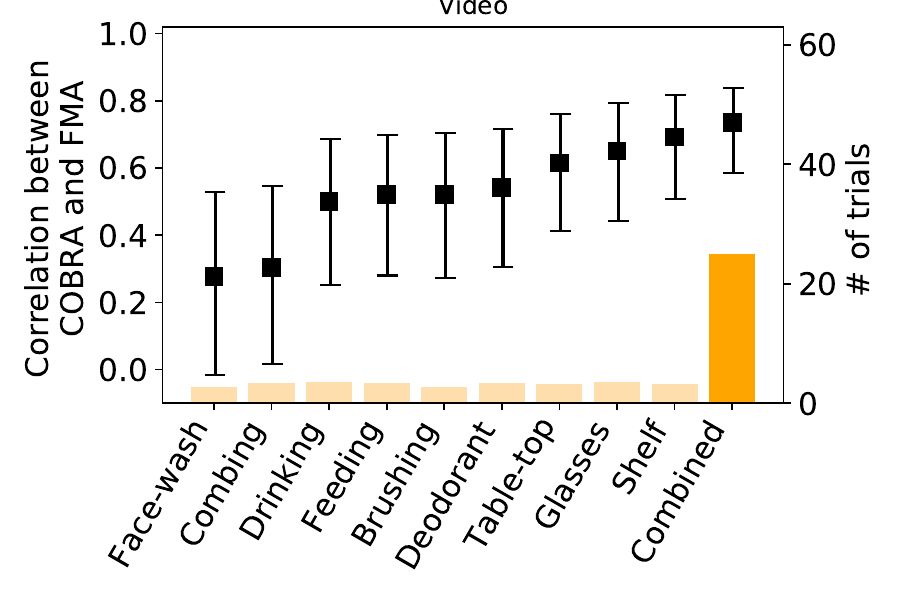}\\
\end{tabular}
\caption{\textbf{Impairment quantification from individual rehabilitation activities.} The graph shows the correlation coefficient (indicated by black markers with 95\% confidence intervals) between the Fugl-Meyer score of stroke patients, and the COBRA score computed from single activities using wearable-sensor (left) or video (right) data. The number of trials available for each activity are indicated by the yellow bars. Simple, more structured activities (Glasses, Shelf, Table-top) have higher correlation than more complicated activities (Face-wash, Feeding, Combing) for both data modalities.  
%Adopting single structured activity (Shelf or Table-top) for calculating COBRA achieves comparable quantification results (right) while using far fewer data. The correlation coefficients are comparable for these two individual activities, despite that the amount of data for calculating COBRA is greatly reduced. 
}\label{fig:COBRA_activities}
\end{figure}

Figure~\ref{fig:COBRA_activities} reports the correlation coefficients between the FMA score and the COBRA score computed using subsets of the data corresponding to individual rehabilitation activities (see Tables~\ref{tab:act_descp_1} and~\ref{tab:act_descp_2} for a detailed description of the activities). Scatterplots of the FMA and COBRA scores for each activity are provided in Figures~\ref{fig:stroke_scatterplots_sensors} and \ref{fig:stroke_scatterplots_video}. For both data modalities, the correlation is higher for more structured activities (moving objects to targets on a table-top or shelf, donning glasses) and is lower for more complex activities (hair-combing, face-washing, teeth-brushing, feeding), which tend to involve more heterogeneous motions across individuals. The correlation coefficient with the FMA score is  lower for the COBRA score computed from individual activities than for the COBRA score that aggregates all activities. The only exception is the table-top task, which is the most regular and structured activity. The corresponding COBRA score computed from wearable-sensor data is very high (0.849, 95\% CI [0.752, 0.910]), which suggests that it may be possible to obtain accurate impairment assessment from a reduced number of data using activities that are highly structured.

\begin{figure}
\begin{tabular}{
 >{\centering\arraybackslash}m{0.3\linewidth} >{\centering\arraybackslash}m{0.3\linewidth}  >{\centering\arraybackslash}m{0.3\linewidth}   }
 Dark + Light & Dark & Light \\
\includegraphics[width=0.9\linewidth]{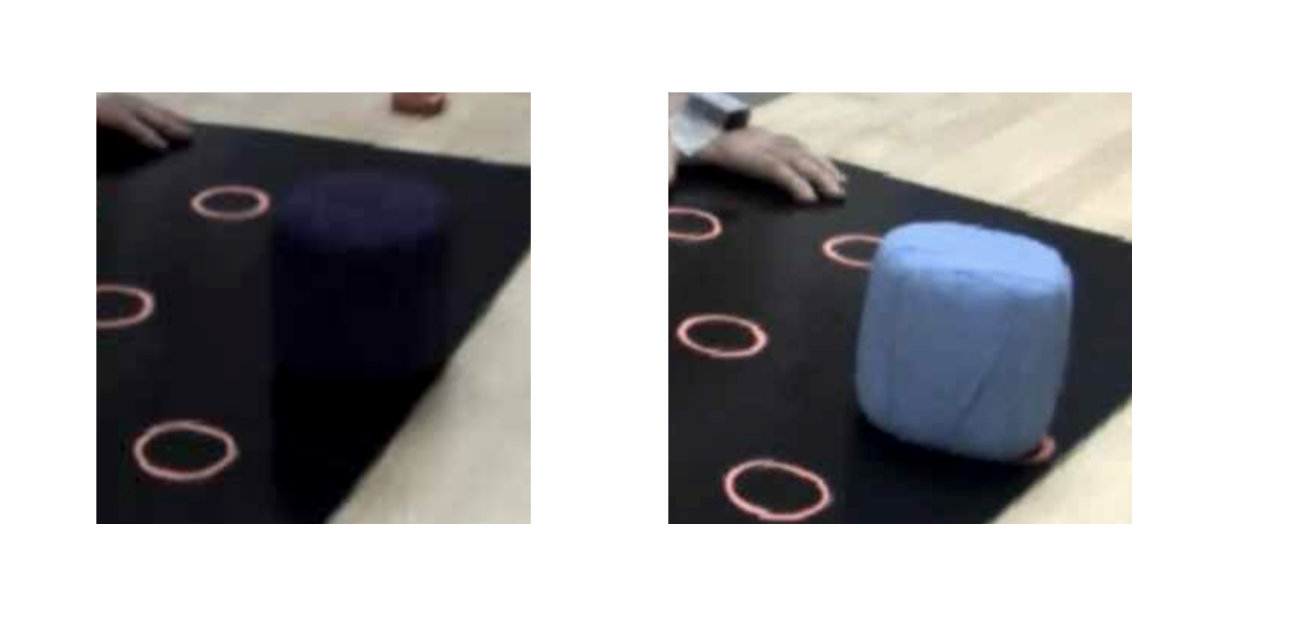}   &\includegraphics[width=0.5\linewidth]{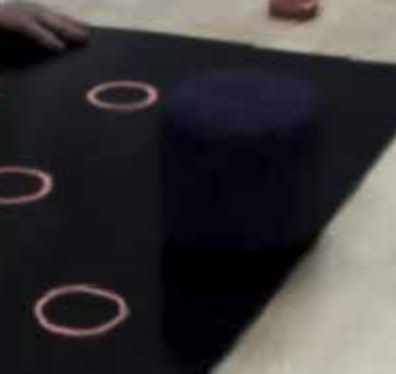} & \includegraphics[width=0.5\linewidth]{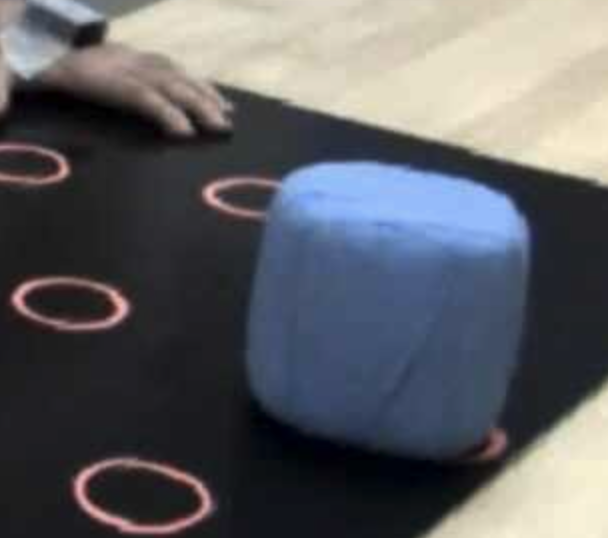} \\
 $\rho$ = 0.615 
 %(95$\%$ CI [0.411,0.760]) 
 &  $\rho$ = 0.679 
% (95$\%$ CI [0.294,0.874]) 
 & $\rho$ = 0.753 
 %(95$\%$ CI [0.560,0.868])
 \\
 {\small 95$\%$ CI [0.411,0.760]} 
 & {\small 95$\%$ CI [0.294,0.874]} 
 &{\small 95$\%$ CI [0.560,0.868]} \\
% \makecell{Correlation coefficient = 0.814\\ (95$\%$ CI [0.700,0.888])} &  \makecell{Correlation coefficient = 0.272 \\ (95$\%$ CI [0.008,0.501])} & \makecell{Correlation coefficient = 0.680 \\(95$\%$ CI [0.506,0.801])}\\
\includegraphics[width=\linewidth]{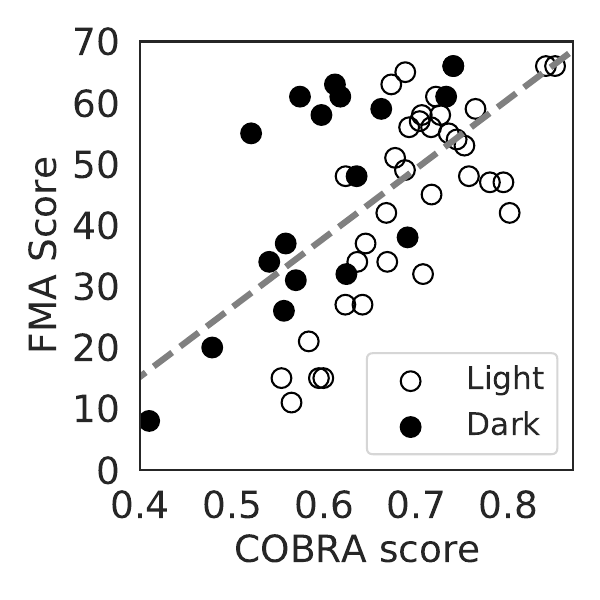} &  \includegraphics[width=\linewidth]{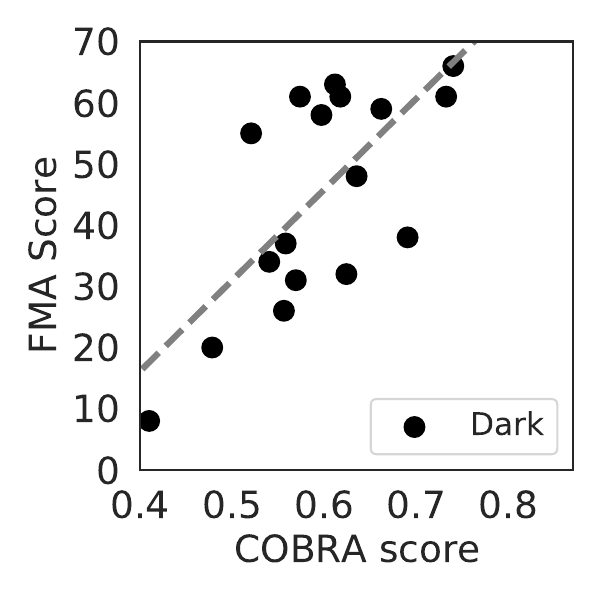} & \includegraphics[width=\linewidth]{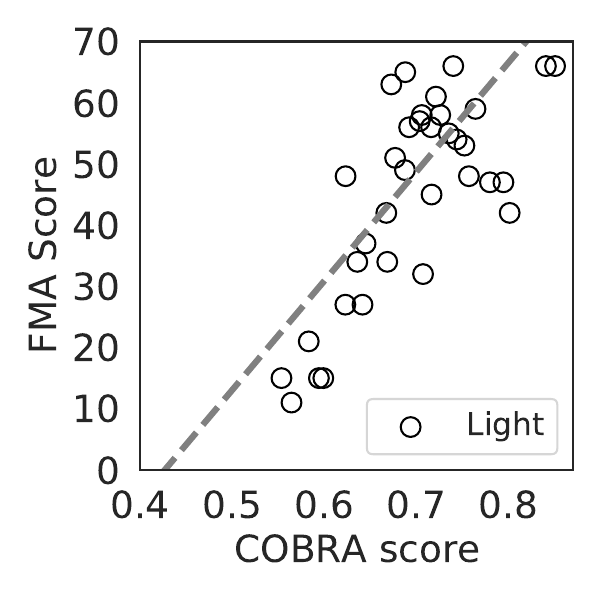} 
\end{tabular}
\caption{\textbf{Object color as a confounding factor for the video-based COBRA score and correction via stratification}. The table-top rehabilitation activity in the stroke impairment quantification task involves dark and light-colored objects (top row). The bottom left scatterplot shows the COBRA score computed only using video data from this activity and the corresponding Fugl-Meyer assessment (FMA) score. The dark objects are difficult to detect, which results in a systematic loss of confidence in the video-based AI model, and hence lower COBRA scores (independently from the FMA score). The bottom middle and right scatterplots show that stratifying according to object color corrects for the confounding factor, improving the correlation coefficient $\rho$ between the COBRA and FMA scores.
}\label{fig:light_dark}
\end{figure}

An important consideration when applying the proposed framework is that extraneous factors may produce a spurious decrease in the confidence of the AI model. Figure~\ref{fig:light_dark} shows that this occurs for the table-top activity, which was carried out with light-colored and dark-colored objects by different subjects. Dark objects are much more difficult to detect in videos, which produces a systematic loss of confidence in the video-based AI model that translates to lower COBRA scores. 
%As illustrated in Figure~\ref{fig:light_dark}, 
This explains why the correlation between the FMA score and the COBRA score is lower for the table-top video data than for the table-top wearable-sensor data, which is unaffected by this confounding factor. 
%for the table-top activity is lower , compared to the wearable-sensor data. 
%This activity was carried out with objects of different colors for different subjects. For subjects with similar levels of impairment, we observe that the COBRA score is systematically lower when the object is dark, which makes it harder to detect (see top of Figure~\ref{fig:light_dark}). 
As depicted in Figure~\ref{fig:light_dark}, we can correct for the confounding factor by stratifying the subjects according to the object color. This increases the COBRA score from 0.615 (95\% CI [0.411, 0.760]) to 0.679 (95\% CI [0.294, 0.874]) for dark objects and 0.756 (95\% CI [0.553, 0.874]) for light objects. For comparison, the correlation of the video-based COBRA score computed from all activities is 0.736 (95\% CI [0.584,0.838]). Figure~\ref{app_fig_blur} shows that image quality can also act as a confounding factor: blurring the video images results in a systematic decrease of the COBRA score, which can also be corrected via stratification. % Our results suggest that video data is more susceptible to such confounding factors than wearable-sensor data. 

\begin{figure}
\begin{tabular}{
 >{\centering\arraybackslash}m{0.3\linewidth} >{\centering\arraybackslash}m{0.3\linewidth}  >{\centering\arraybackslash}m{0.3\linewidth}   }
 \multicolumn{3}{c}{(a) Motion impairment in stroke patients} \vspace{0.2cm}\\
 \multicolumn{3}{c}{ \textbf{Wearable Sensors}} \vspace{0.1cm}\\
 Motion  & Non-motion  & All\\
 $\rho$ = 0.814
 %(95$\%$ CI [0.700,0.888]) 
 & $\rho$ = 0.272 
 %(95$\%$ CI [0.008,0.501]) 
 & $\rho$ = 0.680 
 %(95$\%$ CI [0.506,0.801])
 \\
 {\small 95$\%$ CI [0.700, 0.888]}
 &{\small 95$\%$ CI [0.008, 0.501]}
 & {\small 95$\%$ CI [0.506, 0.801]}\\
 % 0.814 (95$\%$ CI [0.700,0.888]) & 0.272  (95$\%$ CI [0.008,0.501]) & 0.680 (95$\%$ CI [0.506,0.801]) \\
 \includegraphics[width=\linewidth]{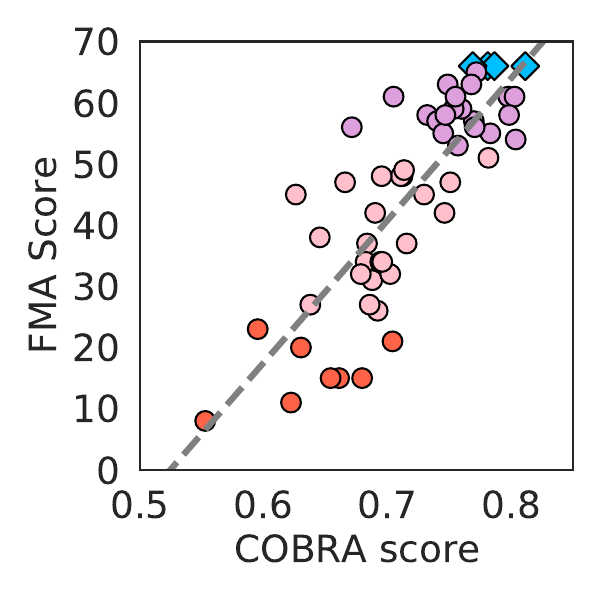} &   \includegraphics[width=\linewidth]{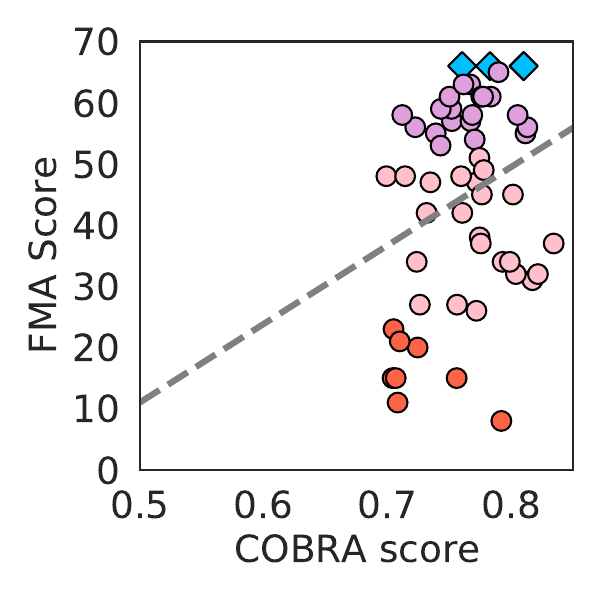} & \includegraphics[width=\linewidth]{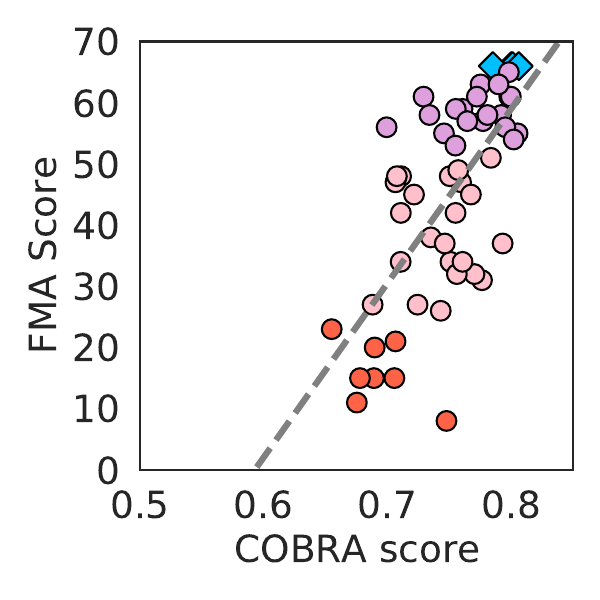} \\
 \multicolumn{3}{c}{ \textbf{Video}} \vspace{0.1cm}\\
 Motion  & Non-motion  & All\\
 $\rho$ = 0.736
 %(95$\%$ CI [0.700,0.888]) 
 & $\rho$ = 0.129 
 %(95$\%$ CI [0.008,0.501]) 
 & $\rho$ = 0.455 
 %(95$\%$ CI [0.506,0.801])
 \\
 {\small 95$\%$ CI [0.584,0.838]}
 &{\small 95$\%$ CI [-0.141,0.382]}
 & {\small 95$\%$ CI [0.216,0.643]}\\
 \includegraphics[width=0.9\linewidth]{figures/video/fig5-video_motion_w_legend.pdf} &   \includegraphics[width=0.9\linewidth]{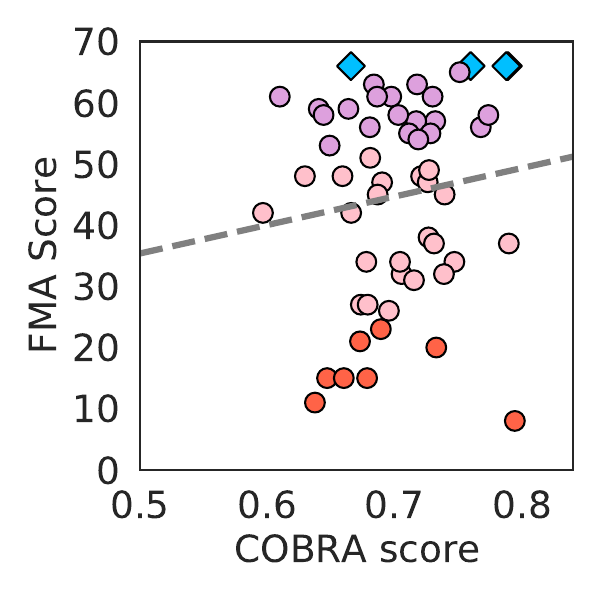} & \includegraphics[width=0.9\linewidth]{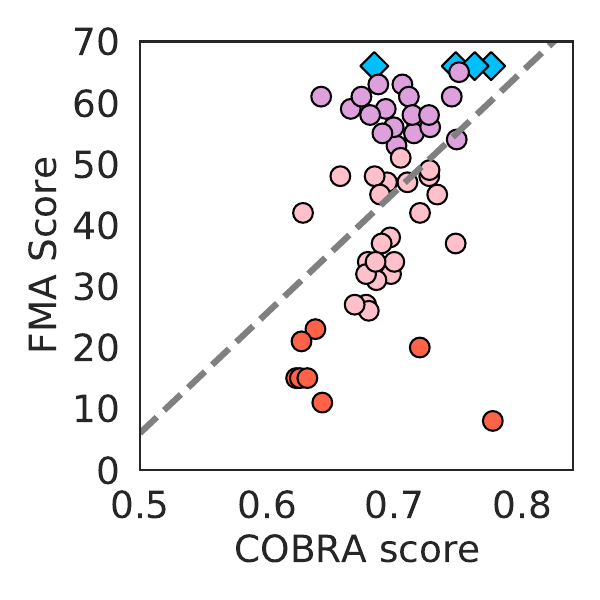} \\
\multicolumn{3}{c}{(b) Severity of knee-osteoarthritis from MRI scans} \vspace{0.2cm}\\
Cartilage & Bone & All tissues\\
 $\rho$ = -0.644
 & $\rho$ = -0.426 
 & $\rho$ = -0.634 
 \\
 {\small 95$\%$ CI [-0.696,-0.585]}
 &{\small 95$\%$ CI [-0.500,-0.346]}
 & {\small 95$\%$ CI [-0.687,-0.575]}\\
\includegraphics[width=\linewidth]{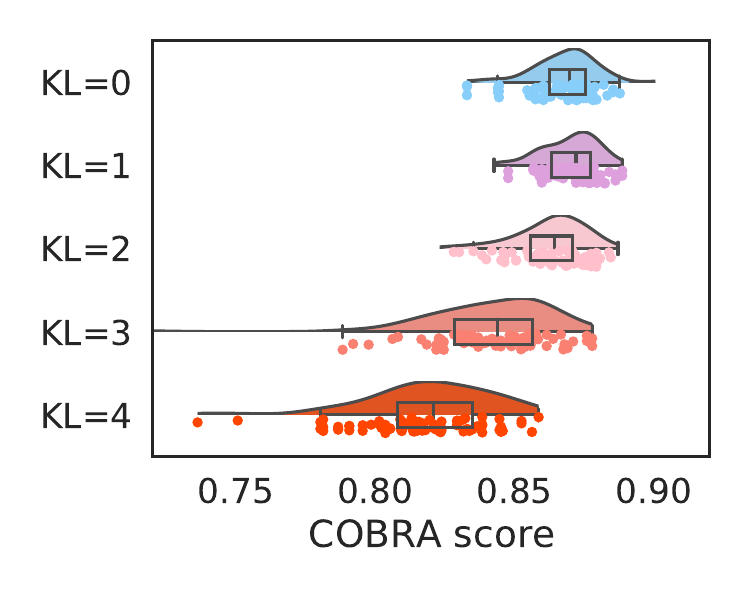} &   \includegraphics[width=\linewidth]{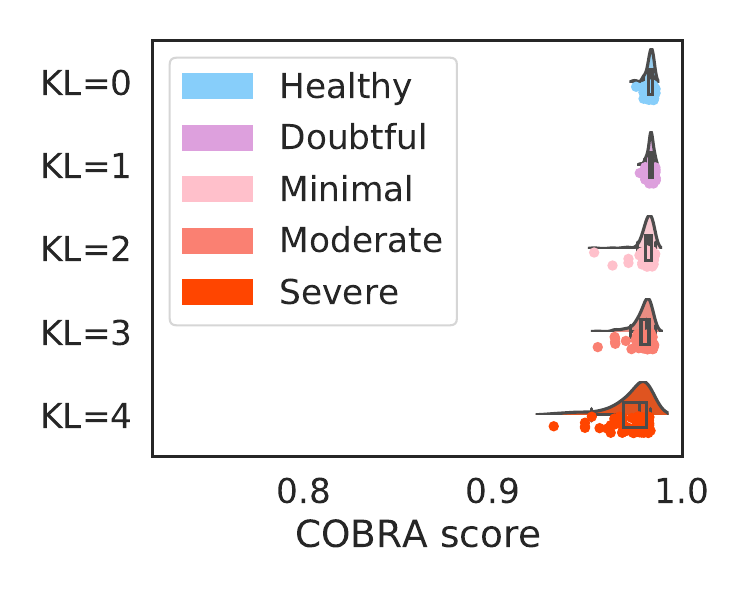} & \includegraphics[width=\linewidth]{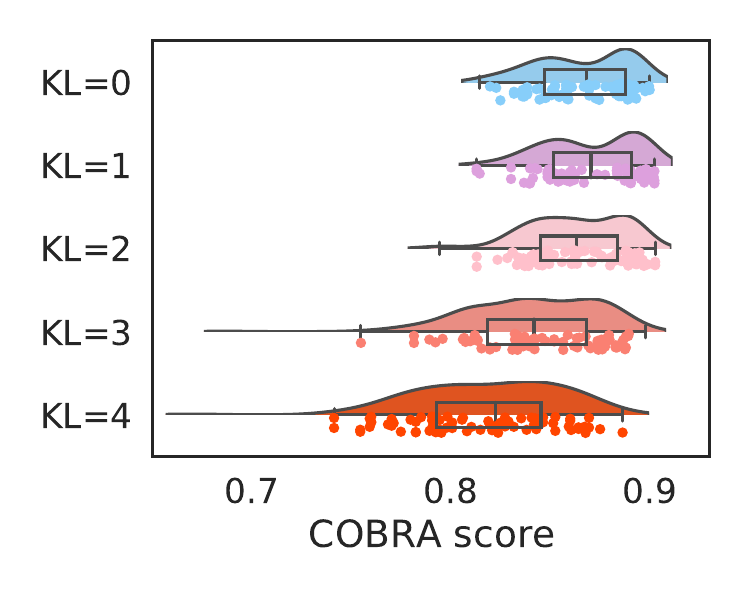} \\
 % Cartilage  &  \multicolumn{2}{c}{\includegraphics[width=0.7\linewidth]{figures/knee/fig3-OAI_cartilage_group_plot.pdf}}\\
 % Bone & \multicolumn{2}{c} {\includegraphics[width=0.7\linewidth]{figures/knee/fig3-OAI_bone_group_plot.pdf}}\\
 % All tissues & \multicolumn{2}{c} {\includegraphics[width=0.7\linewidth]{figures/knee/fig3-OAI_cartilage_and_bone_group_plot.pdf}} \\
\end{tabular}
\caption{\textbf{The COBRA score exploits clinically-relevant structure.} The left column shows scatterplots of clinical assessments (Fugl-Meyer assessment and Kellgren-Lawrence grade for the stroke and knee-osteoarthritis applications respectively) and the proposed COBRA score, computed from data identified as clinically relevant (motion actions for stroke, cartilage tissue for knee osteoarthritis). The middle and right column shows scatterplots for COBRA scores computed from the remaining data, and from all the data respectively. Using clinically relevant data consistently achieves a higher correlation with the clinical assessments.  
%In stroke impairment, focusing on motion-based primitives helps COBRA scores detect stroke-specific movement abnormalities. In knee OA severity, deriving cartilage-specific COBRA scores captures OA-specific disease symptoms around the knee joint area. 
}\label{fig:motion_vs_nomotion}
\end{figure}

The COBRA score is computed as an average of the AI-model confidence for data points identified by the model as corresponding to functional actions that involve motion (\emph{reach}, \emph{reposition}, \emph{transport}), as opposed to functional actions that do not (\emph{rest}, \emph{stabilize}). %\footnote{A description of the different functional movements is provided in Section~\ref{sec:methods_stroke_data}.} 
These data can be considered as \emph{clinically relevant} to impairment quantification associated with motion. Figure~\ref{fig:motion_vs_nomotion} shows that the correlation coefficient between the FMA score and a COBRA score computed from data points identified as  non-motion functional actions is low (in fact, for the video data it is not even statistically significant). It also shows that a COBRA score computed from all actions has a lower correlation with the FMA score than the proposed motion-based COBRA score for both data modalities. 

\subsection{Quantification of Knee-Osteoarthritis Severity}
\label{sec:results_knee}

\begin{table}[t]
\begin{center}
\begin{minipage}{314pt}
\caption{Demographic and clinical characteristics of study participants for the application to quantification of knee-osteoarthritis severity. The mean $\pm$ standard deviation is reported for age.}\label{tab:tab_knee}%
\begin{tabular}{lll}
\toprule
   & Training  & Testing \\
\midrule
Number of individuals & 44   & 435  \\
Age   & 59.2 $\pm$ 8.2    & 62.0 $\pm$ 9.4  \\
Sex  & 20 male, 24 female    & 228 male, 207 female  \\
Race\footnotemark[1]  & 36 W, 7 B, 1 O   & 339 W, 81 B, 5 A, 10 O   \\
\hline
Kellgren-Lawrence grades   & 44 healthy (KL=0) & 
% 57 healthy (KL=0), 58 doubtful (KL=1), 109 minimal (KL=2), 138 moderate (KL=3), 73 severe (KL=4)
\makecell[l]{ 57 healthy (KL=0) \\58 doubtful (KL=1)\\ 109 minimal (KL=2)\\ 138 moderate (KL=3)\\ 73 severe (KL=4)}   
\\
\botrule
\end{tabular}
% \footnotetext{Source: This is an example of table footnote. This is an example of table footnote.}
%\footnotetext[1]{Sex: Male(M), Female(F)}
\footnotetext[1]{Race: White (W), Black (B), Asian (A), Other (O)}
%\footnotetext[3]{Kellgren-Lawrence(KL) grades: 0 for healthy, 1 for doubtful, 2 for minimal, 3 for moderate, 4 for severe  }
\end{minipage}
\end{center}
\end{table}

\textbf{Data}. 
% Our analysis comprised 479 MRI scans of the right knee obtained from the OAI-ZIB dataset~\cite{ambellan2019automated}. This dataset provides manual segmentations of bone and cartilage, which aims to evaluate the severity of knee osteoarthritis (OA), a chronic, degenerative joint disease affecting a significant fraction of the human population. The dataset includes corresponding Kellgren-Lawrence (KL) grading scores~\cite{kohn2016classifications}, which serve as indicators of disease severity on a discrete scale ranging from 0 to 4. Grade 0 signifies the absence of OA, while Grade 4 denotes a severe condition (for more information on the dataset and KL-grading system, refer to Section~\ref{sec:methods_knee_data}). Supplementary figures \ref{app_fig:COBRA_scatterplots} showcase illustrative examples of imaging data associated with distinct KL groups.
The application of the COBRA score to the quantification of knee-osteoarthritis (OA) severity was carried out using the publicly available OAI-ZIB dataset~\cite{ambellan2019automated}. This dataset provides 3D MRI scans of 101 healthy right knees and 378 right knees affected by knee osteoarthritis, a long-term degenerative joint condition. %prevalent in a substantial portion of the global population. 
Each knee is labeled with the corresponding Kellgren-Lawrence (KL) grade~\cite{kohn2016classifications}, retrieved from the NIH Osteoarthritis Initiative collection~\cite{eckstein2012recent}. The KL grade quantifies OA severity on a scale from 0 (healthy) to 4 (severe), as illustrated in Figure~\ref{app_fig:knee_mri_scans}. Each voxel in the MRI scans is labeled to indicate the corresponding tissue (\emph{tibia bone}, \emph{tibia cartilage}, \emph{femur bone}, \emph{femur cartilage} or \emph{background}). % Additional information regarding the dataset and the KL grading system is provided in Section~\ref{sec:methods_knee_data}. %Supplementary figures~\ref{app_fig:knee_mri_scans} present representative examples of imaging data corresponding to varying KL grades.

% Our analysis utilized 479 MRI scans of the right knee, sourced from the OAI-ZIB dataset~\cite{ambellan2019automated}. This dataset provides meticulous segmentations of bone and cartilage, facilitating the assessment of the severity of knee osteoarthritis (OA) – a long-term degenerative joint condition prevalent in a substantial portion of the global population. The dataset also includes corresponding Kellgren-Lawrence (KL) grading scores~\cite{kohn2016classifications}, which function as discrete disease severity markers on a scale from 0 to 4. Grade 0 signifies an absence of OA, while Grade 4 represents an advanced stage of the condition. Additional information regarding the dataset and the KL grading system can be found in Section~\ref{sec:methods_knee_data}. Supplementary figures~\ref{app_fig:knee_mri_scans} present representative examples of imaging data corresponding to varying KL grades.

% AI models were trained to predict the pixel-wise medical segmentation on 44 healthy scans (selected at random). performed by a training cohort consisting of 25 of the 29 healthy individuals (selected at random) separately from the wearable sensor and from the video data. 

\noindent \textbf{Computation of COBRA Score}. An AI segmentation model was trained to predict the tissue at each voxel using a training cohort of 44 healthy individuals (selected at random). A detailed description of this model is provided in Section~\ref{sec:COBRA_knee}. The model was applied to a test cohort consisting of the remaining 57 healthy individuals and the 378 patients with knee OA. Demographic and clinical information about the training and test cohorts is provided in Table~\ref{tab:tab_knee}. The COBRA score was calculated as the average of the model confidence for data points identified by the models as corresponding to cartilage tissue (\emph{tibia cartilage} and \emph{femur cartilage}), as described in Section~\ref{sec:COBRA_knee}. %Detailed segmentation performances and further implementation details can be found in Supplementary Section~\ref{sec_app:Knee_OA}

% We develop a semantic segmentation model for understanding a healthy knee structure and for which we utilized 44 healthy scans. Specifically, the model was trained to perform an auxiliary task of segmenting the femur bone, femur cartilage, tibia bone, and tibia cartilage for each pixel in the image. Our segmentation model, trained on healthy data, achieved a dice coefficient of 0.916 for held-out healthy knees, indicating its ability to accurately capture normal knee structure. Detailed segmentation performances and further implementation details can be found in Supplementary Section~\ref{sec_app:Knee_OA}. Subsequently, using the segmentation model, we proceeded to assess knee OA severity by computing COBRA scores for the remaining 378 damaged knee images as well as 57 held-out healthy knee images.  

\noindent \textbf{Evaluation}. The COBRA score was evaluated by computing its Pearson correlation coefficient with the Kellgren-Lawrence (KL) grading scores~\cite{kohn2016classifications} on the test cohort of 378 patients with knee OA and 57 healthy subjects (n=435), which equals -0.644 (95\% CI [-0.696, -0.585]). There is therefore a significant inverse correlation between the scores, indicating that the COBRA score quantifies knee OA severity. Figure \ref{fig:COBRA_scatterplots}(b) shows scatterplots and density plots of the COBRA scores corresponding to different KL grades.
% , the model's confidence decreases as the severity of knee joint abnormality increases (higher KL values). We observe an overall Pearson correlation of -0.644 (95\% CI [-0.696, -0.585]) between our derived COBRA scores and the KL-grades for the knee.

% Similar to the stroke dataset, in this case as well, we rely on clinical insights to calculate our COBRA score. Clinical evidence indicates that knee OA is often characterized by progressive loss of articular cartilage and narrowing of the joint space~\cite{hsu2018knee, brody2015knee}, primarily affecting the cartilage area. While changes in the bone and the formation of bone spurs (osteophytes) can occur in certain cases, they are less common. Consequently, in the context of knee OA, cartilage plays a more significant role than bone. Figure \ref{fig:COBRA_scatterplots} demonstrates the effectiveness of COBRA scores derived from cartilage segmentation in identifying abnormal knees, exhibiting a correlation coefficient of -0.644 (CI [-0.696, -0.585]). In comparison, the correlation coefficient for bone-derived scores is -0.441 (CI [-0.537, -0.334]).

The COBRA score is computed as an average of the AI-model confidence for voxels identified by the model as corresponding to cartilage, as opposed to bone tissue. %\footnote{A description of the different functional movements is provided in Section~\ref{sec:methods_stroke_data}.} 
These data can be considered as \emph{clinically relevant} because knee OA produces gradual degradation of articular cartilage (bone alterations and osteophyte formation may also occur, but are less frequent)~\cite{hsu2018knee, brody2015knee}. Figure~\ref{fig:motion_vs_nomotion} shows that the magnitude of the correlation coefficient between the KL score and a COBRA score is significantly lower than for cartilage. The magnitude of the correlation coefficient for the COBRA score computed from all voxels is only slightly lower than that of the proposed motion-based COBRA score, indicating that including bone is not very detrimental.

\section{Discussion}\label{sec:discussion}

In this work we introduce the COBRA score, a data-driven anomaly-detection framework for automatic quantification of impairment and disease severity. We show its utility for clinically relevant quantification in two different medical conditions (stroke and knee osteoarthritis) and in three different data modalities (wearable sensors, video and MRI). The framework is suitable for applications where it is challenging to gather large-scale databases of patients with different degrees of impairment or severity, as it only requires data from a healthy cohort of moderate size. 

From a methodological perspective, our results suggest that fine-grained annotations describing clinically relevant attributes can be useful \emph{even if they are only available for healthy subjects}. We hypothesize that AI models trained with such annotations can be leveraged in different ways beyond the proposed approach. To illustrate this, an alternative anomaly-detection procedure that does not utilize model confidence is included in Section~\ref{sec:distance_based}.

%calculating the distance between abnormal and healthy samples' representation~\cite{heusel2017gans}, in the supplementary section \ref{secA1}.
%When the models are presented with data where the attribute is affected by a medical condition (i.e. stroke-induced motion impairment, or cartilage damage), the model confidence tends to decrease proportionally to the severity of the condition. Our results demonstrate that aggregating this confidence yields a data-driven metric (the COBRA score), which has a strong correlation with independent clinical assessments.   

Our study identifies a key consideration when applying the proposed framework: confounding factors unrelated to the medical condition of interest (e.g. object color or blurriness in a video) can influence the confidence of the AI models, and therefore distort the COBRA score. This is an instance of a general challenge inherent to the use of deep neural networks: these models are so flexible that they can easily learn spurious structure in high-dimensional data~\cite{geirhos2020shortcut,zhang2021understanding}. Our results suggest that the influence of confounding factors can be mitigated by gathering a sufficiently diverse training set (e.g. including diverse rehabilitation activities in the case of stroke-induced impairment) and that it is possible to explicitly correct for known confounding factors via stratification. Nevertheless, automatic identification and control of confounders is an important topic for future research.  

\section{Methods}\label{sec:methods}

In this section we describe a general framework to estimate impairment and disease severity using AI models trained only on data from healthy subjects. We frame this as an anomaly detection and quantification problem, where the goal is to identify subjects that deviate from the healthy population, and to quantify the extent of this deviation. Section~\ref{sec:COBRA_method} describes the proposed framework, and Sections~\ref{sec:COBRA_stroke}  and~\ref{sec:COBRA_knee} describe how we apply it to quantify impairment in stroke patients and severity of knee osteoarthritis respectively.

\subsection{Confidence-based Characterization of Anomalies} 
\label{sec:COBRA_method}

%The proposed COnfidence-Based chaRacterization of Anomalies (COBRA) score is based on an AI model trained to perform a clinically-relevant classification task on a training cohort of healthy subjects.
The proposed COnfidence-Based chaRacterization of Anomalies (COBRA) framework utilizes a model trained to perform an AI task only on healthy patients. Intuitively, if the model has low confidence when performing the task on a new subject, this indicates that the subject deviates from the healthy population. In order to ensure that this deviation is due to a certain type of impairment or disease, it is crucial to choose an appropriate AI task. For quantification of stroke-induced impairment, we predict the functional actions carried out by the subject from wearable sensor or video data. For the application to knee osteoarthritis, we predict the tissue present in each voxel of a 3D MRI scan.  

Let us assume that we have access to a training cohort of $N_{\operatorname{train}}$ healthy subjects, and that each of them is associated with a set of annotated data relevant to the medical condition of interest:
\begin{align}
\mathcal{T}_i := \left\{ \left( x_1^{[i]},y_1^{[i]}\right),\ldots,\left(x_{M_i}^{[i]},y_{M_i}^{[i]}\right) \right\}, \quad 1 \leq i \leq N_{\operatorname{train}}.
\end{align}
Here $x_j^{[i]} \in \mathbb{R}^{L}$ denotes the $j$th data point associated with the $i$th subject, and $M_i$ is the number of data available for that subject. %In our applications of interest, the data are acquired using wearable sensors, video or magnetic-resonance imaging (MRI). 
The label $\smash{y_j^{[i]}} \in \left\{1,\ldots,K\right\}$ assigns $\smash{x_j^{[i]}}$ to one of $K$ predefined classes. For the stroke application, the label encodes the functional action carried out by the subject at a certain time. The corresponding data point is a segment of wearable-sensor or video data. For the knee-osteoarthritis application, the label encodes the type of tissue at a certain position in the knee, and the corresponding data are surrounding MRI voxels. 

The training dataset
 \begin{align}
\mathcal{S}_{\operatorname{train}}:= \left\{ \mathcal{T}_1, \ldots,\mathcal{T}_{N_{\operatorname{train}}} \right\}
 \end{align}
is used to train an AI model $f:\mathbb{R}^{L}\rightarrow [0,1]^{K}$ to predict the labels from the data. The input to the model is an $L$-dimensional data point and the output is a $K$ dimensional vector
%\left\{1,\ldots,K\right\}$
\begin{align}
p^{[i]}_j := f(x_j^{[i]}), \quad 1\leq i \leq N_{\operatorname{train}}, \; 1\leq i \leq M_i, 
\end{align}
where the $k$th entry is an estimate of the probability that the data point belongs to the $k$th class. In our applications of interest, the models are deep neural networks, described in detail in Sections~\ref{sec:COBRA_stroke} and \ref{sec:COBRA_knee}. 
%trained by minimizing the cross entropy between the estimated probabilities and the labels over the training set~\cite{goodfellow2016deep}. 
Crucially, if the dataset associated with each subject is large, then the total number of training examples
\begin{align}
M_{\operatorname{train}} := \sum_{i=1}^{N_{\operatorname{train}}} M_i
\end{align}
is orders of magnitude larger than the number of training subjects $N_{\operatorname{train}}$. This enables us to train deep-learning models using relatively small training cohorts.

% \begin{align}
% \hat{y}^{[i]}_j & := \arg \max_{1 \leq k \leq K} p^{[i]}_j[k], \\
% c^{[i]}_j & := \max_{1 \leq k \leq K} p^{[i]}_j[k] 
% \end{align}

Let 
%\begin{align}
$X_{\operatorname{test}} := \left\{x_1^{\operatorname{test}},\ldots,x_{M_{\operatorname{test}}}^{\operatorname{test}} \right\}$ 
%\end{align}
denote a dataset associated with a test subject. We can obtain probabilities corresponding to each test data point by applying the trained AI model,
\begin{align}
p^{\operatorname{test}}_j := f(x_j^{\operatorname{test}}), \quad 1\leq j \leq M_{\operatorname{test}}. 
\end{align}
This yields a prediction of the class associated with each example 
\begin{align}
\hat{y}^{\operatorname{test}}_j & := \arg \max_{1 \leq k \leq K} p^{\operatorname{test}}_j[k],
\end{align}
where $p^{\operatorname{test}}_j[k]$ denotes the $k$th entry of $p^{\operatorname{test}}_j$. The estimated probability that the data point belongs to the predicted class is commonly known as the \emph{confidence} of the model (see e.g. \cite{guo2017calibration}),
\begin{align}
c^{\operatorname{test}}_j & := \max_{1 \leq k \leq K} p^{\operatorname{test}}_j[k],
\end{align}
because it can be interpreted as an estimate of the probability that the model prediction is correct. 

Several existing works propose to use confidence values to perform anomaly detection~\cite{hendrycks2016baseline,chen2020robust,hsu2020generalized,vyas2018out,mohseni2020self,devries2018learning}. Intuitively, if a model is well trained (and there is no inherent uncertainty in the training labels~\cite{liu2022deep}), it should be able to confidently classify new examples. Therefore low model confidence is evidence that the data point may be anomalous, in the sense that it deviates from the training distribution. Our proposed framework builds upon this idea, incorporating two novel elements. First, multiple data points are aggregated to perform subject-level anomaly detection. As illustrated by Figure~\ref{fig:COBRA_histograms}, this is critical to achieve accurate anomaly detection in our applications of interest, because the individual confidences are very noisy. Second, we determine which of the classes are most clinically-relevant, and restrict our attention to data points predicted to belong to those classes. As reported in Figure~\ref{fig:motion_vs_nomotion}, for the stroke application this provides a substantial improvement over using all the data.

Let $\mathcal{R} \subseteq \left\{1,\ldots,K\right\}$ denote the subset of clinically-relevant classes, and
\begin{align}
\mathcal{J}_{\operatorname{relevant}} & := \left\{ j :  \hat{y}^{\operatorname{test}}_j \in \mathcal{R} \right\}
\end{align}
the subset of test data predicted to belong to those classes. We define the COBRA score as the arithmetic average of the confidences associated with the data in $\mathcal{J}_{\operatorname{relevant}}$, 
\begin{align}
\operatorname{COBRA}(X_{\operatorname{test}}) & := \frac{1}{ \lvert \mathcal{J}_{\operatorname{relevant}}
\rvert} \sum_{j\in \mathcal{J}_{\operatorname{relevant}}} c^{\operatorname{test}}_j.
\end{align}
The lower the COBRA score, the less confident the AI model is on average when performing the task on the test subject, which indicates a greater degree of impairment or disease. %As reported in Section~\ref{sec:results}, the COBRA score quantifies such deviations in a way that is consistent with clinical assessments for our two applications of interest.

\subsection{Estimation of Stroke-Related Motor Impairment}
%\label{sec:COBRA_sensors}
\label{sec:COBRA_stroke}
In order to apply the COBRA framework to automatic impairment quantification in stroke patients, we propose to utilize auxiliary AI models trained to predict the functional primitive carried out by the subjects' paretic upper extremity (UE) while performing rehabilitation activities. The $K:=5$ primitive classes are \emph{reach}, \emph{reposition}, \emph{transport}, \emph{stabilize}, and \emph{rest}.  UE motor impairment affects the three functional primitives involving motion 
\begin{align}
\mathcal{R} := \left\{\text{\emph{transport}, \emph{reposition},  \emph{reach}} \right\},
\end{align} 
rendering them systematically different to those of healthy individuals. Our hypothesis is that this causes AI models trained on healthy subjects to lose confidence when they are applied to stroke patients, and that the loss of confidence is indicative of the degree of impairment.  %Section~\ref{sec:results_stroke} shows that 
In the remainder of this section we describe the AI models that we use to test this hypothesis for two different data modalities, wearable sensors and video.

\subsubsection{Wearable sensors} \label{subsec:model_arch_wearable_sensors}
The wearable-sensor data is a 77-dimensional time series, recorded at 100 Hz using nine inertial measurement units (IMUs) attached to the upper body~\cite{kaku2022strokerehab}. The data correspond to kinematic features of 3D linear accelerations, 3D quaternions, joint angles from the upper body, and a binary value that indicates the side (left or right) performing the motion. % As an additional feature, we included side of hand (left or right) in a one-hot encoded value, increasing the dimension of the final feature vector to 77. 
%Each entry of the time series is annotated to indicate the functional action carried out by the subject at that time. %At each time point, the 77-length feature vector is assigned with a primitive label. % An action sequence consists of primitives over time. The clinically meaningful auxiliary task of action segmentation is to predict the primitive from the feature vector for each time point.
In order to identify functional primitives from these data, 
%\textbf{StrokeRehab wearable-sensor data.} To capture healthy movement patterns, 
we utilized a Multi-Stage Temporal Convolutional Network (MS-TCN)~\cite{farha2019ms}. 

MS-TCN is a state-of-the-art deep-learning model for action segmentation consisting of four convolutional stages, each composed of 10 layers of dilated residual convolutions with 64 output channels. A softmax layer at the end of the network produces the final output, which is a 5-dimensional vector indicating the probability that each entry in the time series corresponds to each functional primitive. %In the notation of Section~\ref{sec:COBRA_method}, the input dimensionality $L$ equals $ 77 \times ???? = $. 
The model was trained on the healthy training cohort using the weighted cross-entropy loss function proposed in~\cite{ishikawa2021alleviating}. This cost function was minimized for 50 epochs using the Adam optimizer~\cite{kingma2014adam} with a learning rate of $5\cdot 10^{-3}$ (selected via cross-validation). The accuracy and precision of the resulting model are shown in Appendix~\ref{app_tab_test_performance_sensor_all_subjects}. %The training data corresponding to each subject consists of ???? trials, each of length between ???? and ????, for each of the nine rehabilitation activities.  

\subsubsection{Video}
The video data were acquired with two high-speed (60 Hz), high-definition (1088 × 704 resolution) cameras (Ninox, Noraxon) positioned orthogonally <2 m from the subject. The cameras have a focal length of f4.0 mm and a large viewing window (length: 2.5 m, width: 2.5 m). The videos were then downsampled to a resolution of 256 $\times$ 256 to enable efficient processing. To perform functional primitive identification from these data, we utilized the X3D model~\cite{feichtenhofer2020x3d}, a 3D convolutional neural network designed for primitive classification from video data. %The X3D model is a family of efficient video networks that progressively expand a tiny 2D image classification architecture along multiple network axes, in space, time, width and depth. 
%The X3D model achieves state-of-the-art performance while requiring 4.8x and 5.5x fewer multiply-adds and parameters for similar accuracy as previous work. 
The model was pretrained on the Kinetic dataset~\cite{kay2017kinetics}, where the labels are high-level activities such as running, climbing, sitting, etc. 

After pretraining, the X3D model was fine-tuned to perform classification of functional primitives on the  rehabilitation activities performed by the healthy training cohort. The input to the model are video segments with duration two seconds, and the output is the estimated probability that the central frame corresponds to each of the five functional primitives. Model fine-tuning was carried out by minimizing the cross entropy between these probabilities and the functional primitive labels via stochastic gradient descent with a base learning rate of 0.01 and a cosine learning rate policy. The accuracy and precision of the resulting model on held-out subjects are reported in Section~\ref{sec_app:stroke_impairment}.

%Since the StrokeRehab dataset consists of subtle, sub-second actions, we fine-tuned the X3D model on the training set of StrokeRehab~\cite{kaku2022strokerehab}. For finetuning the model, we use video sequences as input and try to identify the primitive happening in the center frame of the videos. The fine-tuned model was then used to predict the action of the center frame in the 2 seconds clip. 

% .{~\color{red} todo the more details}

\subsection{Estimation of Knee-Osteoarthritis Severity}
\label{sec:COBRA_knee}
In order to apply the COBRA framework to automatic quantification of knee-osteoarthritis severity we propose to utilize an auxiliary AI model trained to predict the type of tissue in each voxel of a 3D MRI scan. The $K:=5$ classes for this classification problem are \emph{femur bone}, \emph{femur cartilage}, \emph{tibia bone}, \emph{tibia cartilage} and \emph{background} (indicating absence of tissue). Knee osteoarthritis deforms cartilage structure, so the clinically-relevant labels are chosen to be
\begin{align}
\mathcal{R} := \left\{\emph{femur cartilage},\text{\emph{tibia cartilage}} \right\}.
\end{align} 
Our hypothesis is that the systematic difference in cartilage structure causes AI models trained on healthy knees to lose confidence when applied to diseased knees, and that the loss of confidence is indicative of disease severity.

%leverages multi-planar augmentation, training multiple 2D U-Net models on images that have been randomly sampled from multiple planes. This forces the network to learn to segment the input seen from different views. The model obtained a top-5 position at the 2018 Medical Segmentation Decathlon\footnote{\url{http://medicaldecathlon.com/}} despite its simplicity and computational efficiency. 
In order to predict tissue type  we applied a Multi-Planar U-Net~\cite{perslev2019one}. The model processes the input 3D MRI scan from different views using a version of the 2D U-Net architecture~\cite{ronneberger2015u}. The output from the different views are then averaged to produce a probability estimate at each 3D voxel. During training, random elastic deformations (RED) are randomly applied to a third of the images in each batch to improve generalization~\cite{perslev2019one}. 

The model was trained by minimizing the cross entropy loss between the estimated probabilities and the 3D voxel-wise labels corresponding to 37 of the 44 healthy individuals in the training cohort. The remaining 7 individuals were used as a validation set. In the cost function, images augmented via RED were downweighted by a factor of $1/3$. The Adam optimizer was used for minimization, with an initial learning rate of $5\cdot 10^{-5}$ that was reduced by $10\%$ after two consecutive epochs without improvement in the validation Dice score. A criterion based on the validation Dice score (excluding background) was used to perform early stopping. Additional hyperparameters are listed in Table S.1 of~\cite{perslev2019one}. The accuracy and precision of the resulting model are reported in Section~\ref{app_tab_test_performance_knee}.

\section*{Declarations}

\begin{itemize}
\item Authors' contributions: H.S. and C.F.G. conceived the project. B.Y., A.K., K.L. and A.P. designed, implemented and evaluated the methodology with guidance from R.R., H.S. and C.F.G. A.P., E.F, A.V., and N.P. quality-checked the data and their annotations. B.Y., A.K., K.L., A.P., R.R., H.S. and C.F.G. wrote the paper with input from all authors. 

\item Competing interests
The authors declare no competing interests.

\item Acknowledgements
This work was supported by NIH grant R01 LM013316, Alzheimer´s Association grant AARG-NTF-21-848627, and NSF grant NRT-1922658.
% \item Ethics approval 
% \item Consent to participate
% \item Consent for publication
\item Data availability: Data to reproduce all results are available at \url{https://github.com/fishneck/COBRA/tree/main/data}.
%StrokeRehab wearable-sensor data and OAI-ZIB MRI data are publicly available at and \url{https://simtk.org/projects/primseq} and \url{https://pubdata.zib.de/}. StrokeRehab video data are stored in an institutional repository and will be shared upon request to the corresponding author.
\item Code availability: Code to reproduce all results is available at \url{https://github.com/fishneck/COBRA}.
\end{itemize}

% \noindent
% If any of the sections are not relevant to your manuscript, please include the heading and write `Not applicable' for that section. 

% \bigskip

% Springer journals and proceedings: \url{https://www.springer.com/gp/editorial-policies}

% Nature Portfolio journals: \url{https://www.nature.com/nature-research/editorial-policies}

% \textit{Scientific Reports}: \url{https://www.nature.com/srep/journal-policies/editorial-policies}

% BMC journals: \url{https://www.biomedcentral.com/getpublished/editorial-policies}
% \end{flushleft}
\bibliography{sn-bibliography}% common bib file

\newpage
\begin{appendices}

\section{Additional Dataset Information}

\begin{figure}
\begin{tabular}
{>{\centering\arraybackslash}m{0.2\linewidth} > {\centering\arraybackslash}m{0.12\linewidth} > {\centering\arraybackslash}m{0.12\linewidth} >{\centering\arraybackslash}m{0.12\linewidth} > {\centering\arraybackslash}m{0.12\linewidth} > {\centering\arraybackslash}m{0.12\linewidth}}

 & KL=0 & KL=1 & KL=2 & KL=3 & KL=4 \\
Knee MRI & \includegraphics[width=\linewidth]{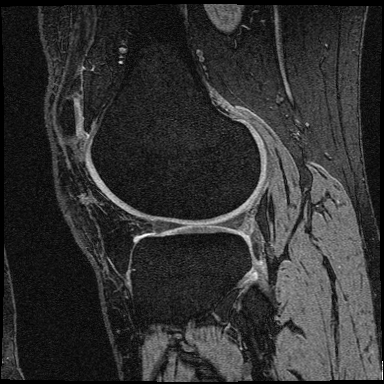} &\includegraphics[width=\linewidth]{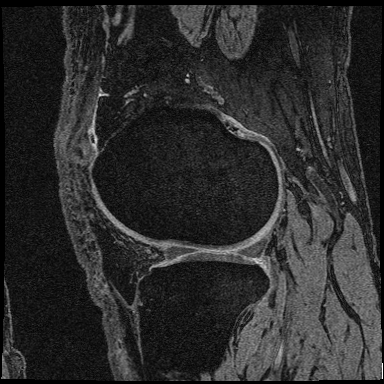} &\includegraphics[width=\linewidth]{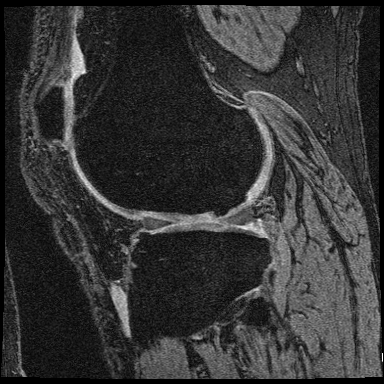} & \includegraphics[width=\linewidth]{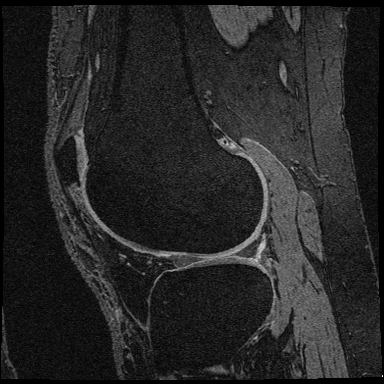} & \includegraphics[width=\linewidth]{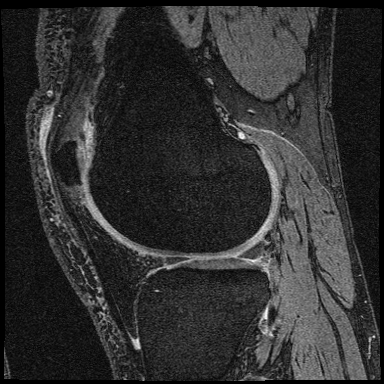}\\
Segmentation & \includegraphics[width=\linewidth]{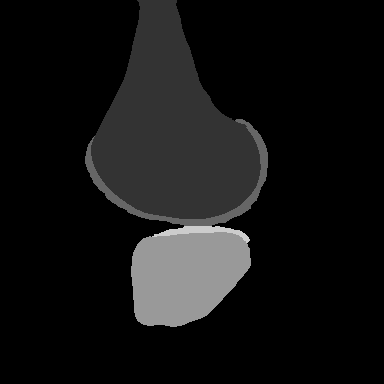} &\includegraphics[width=\linewidth]{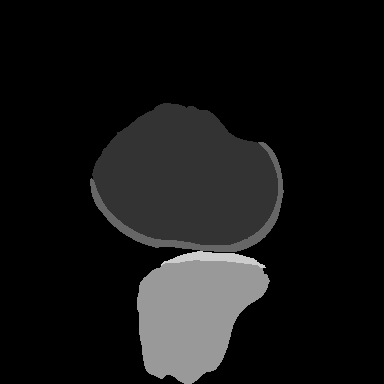} &\includegraphics[width=\linewidth]{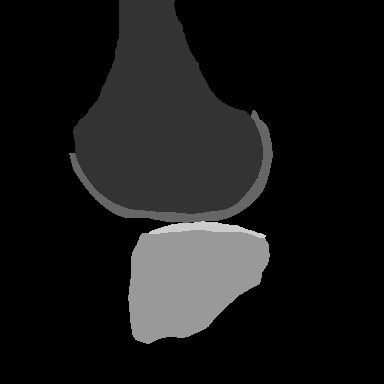} & \includegraphics[width=\linewidth]{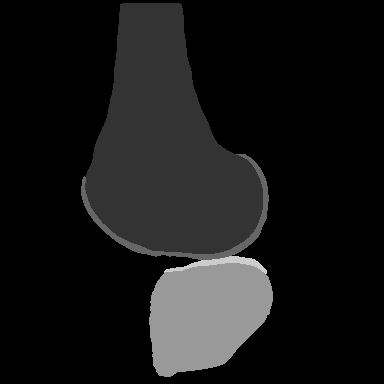} & \includegraphics[width=\linewidth]{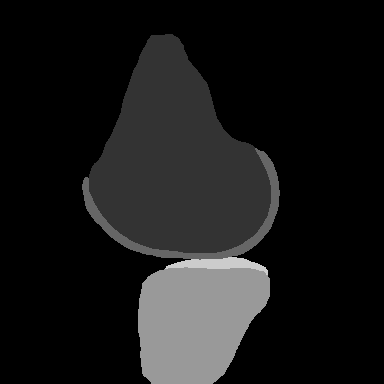}\\
\multicolumn{6}{c}{\includegraphics[width=0.9\linewidth]{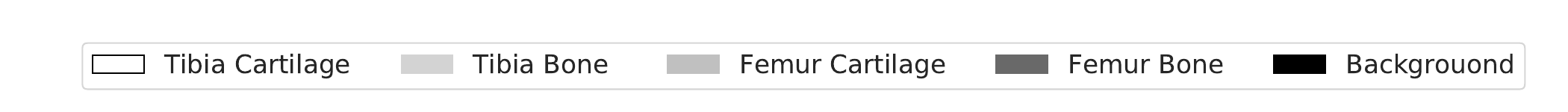}} 
\end{tabular}
\caption{Knee MRI images for subjects with different Kellgren-Lawrence (KL) grades (top) and corresponding segmentation annotations (bottom) indicating the tissue in each pixel.
}\label{app_fig:knee_mri_scans}
\end{figure}

% \subsection{StrokeRehab} \label{sec:app_stroke_dataset}
Tables~\ref{tab:act_descp_1} and \ref{tab:act_descp_2} provide a detailed description of the rehabilitation activities carried out by the subjects in the dataset used for quantification of stroke-induced impairment. Figure~\ref{app_fig:knee_mri_scans} shows examples of the MRI scans used for quantification of knee-osteoarthritis severity.
%In the StrokeRehab dataset, participants were asked to perform 9 activities. They are Brushing, Combing, Deodorant, Drinking, Face-wash, Feeding, Glasses, Shelf and Table-top (ordered alphabetically). Activity descriptions can be found in Tables~\ref{tab:act_descp_1} and \ref{tab:act_descp_2}.  

% \newcolumntype{C}[1]{>{\centering\arraybackslash}p{#1}}
\begin{table*}[!htbp]
\caption{Description of the activities performed by the subjects in the dataset used for quantification of stroke-induced impairment (1/2).}
\resizebox{\textwidth}{!}{%
\centering
\renewcommand{\arraystretch}{1.5} % Default value: 1
\begin{tabular}{|p{2cm}|p{4.5cm}|p{2cm}|p{5cm}|}
\hline
\multicolumn{1}{|c|}{Activity} & \multicolumn{1}{c|}{Workspace} & \multicolumn{1}{c|}{\begin{tabular}[c]{@{}c@{}}Target\\ object(s)\end{tabular}} & \multicolumn{1}{c|}{Instructions} \\ 
\hline
Face-wash & Sink with a small tub (32.3 x 24.1 x 2.5 cm³) in it and two folded washcloths on either side of the countertop, 30 cm from edge closest to patient & Washcloths, faucet handle, and tub & Fill tub with water, dip washcloth on the right side into water, wring it, wiping each side of their face with wet washcloth, place it back on countertop. Use washcloth on the left side to dry face, place it back on countertop \\ \hline
Deodorant & Tabletop with deodorant placed at midline, 25 cm from edge closest to patient & Deodorant (solid twist-base) & Remove cap, twist base a few times, apply deodorant, replace cap, untwist the base, put deodorant on table \\ \hline
Combing & Tabletop with comb placed at midline, 25 cm from edge closest to patient & Comb & Pick up comb and comb both sides of head \\ \hline
Glasses & Tabletop with glasses placed at midline, 25 cm from edge closest to patient & Pair of glasses & Wear glasses, return hands to table, remove glasses and place on table\\ \hline
Feeding & Table top with a standard-size paper plate (21.6 cm diameter) placed at midline, 2 cm from edge, utensils placed 3 cm from edge, 5 cm from either side of plate, a baggie with a slice of bread placed 25 cm from edge, 23 cm left of midline, and a margarine packet placed 32 cm from edge, 17 cm right of midline & Paper plate, fork, knife, re-sealable sandwich baggie, slice of bread, single-serve margarine container & Remove bread from plastic bag and put it on plate, open margarine pack and spread it on bread, cut bread into four pieces, cut off and eat a small bite-sized piece \\ \hline
\end{tabular}}

\label{tab:act_descp_1}
\end{table*}

\begin{table*}[t]
\caption{Description of the activities performed by the subjects in the dataset used for quantification of stroke-induced impairment (2/2).}
\resizebox{\textwidth}{!}{%
\centering
\renewcommand{\arraystretch}{1.5}
\begin{tabular}{|p{1.75cm}|p{4.5cm}|p{2.5cm}|p{5cm}|}
\hline
\multicolumn{1}{|c|}{Activity} & \multicolumn{1}{c|}{Workspace} & \multicolumn{1}{c|}{\begin{tabular}[c]{@{}c@{}}Target\\ object(s)\end{tabular}} & \multicolumn{1}{c|}{Instructions} \\ \hline
Drinking & Tabletop with water bottle and paper cup 18 cm to the left and right of midline, 25 cm from edge closest to patient & Water bottle (12 oz),
paper cup (4 oz) & Open water bottle, pour water into cup, take a sip of water, place cup on table, and replace cap on bottle \\ \hline
Brushing & Sink with toothpaste and toothbrush on either side of the countertop, 30 cm from edge closest to patient & Travel-sized toothpaste, toothbrush with built-up foam grip, faucet handle & Wet toothbrush, apply toothpaste to toothbrush, replace cap on toothpaste tube, brush teeth, rinse toothbrush and mouth, place toothbrush back on countertop \\ \hline
Table-top & Horizontal circular array (48.5 cm diameter) of 8  targets (5 cm diameter) & Toilet paper roll wrapped in self-adhesive wrap  & Move the roll between the center and each outer target, resting between each motion and at the end \\ \hline
Shelf & Shelf with two levels (33 cm and 53 cm) with 3 targets on both levels (22.5 cm, 45 cm, and 67.5 cm away from the left-most edge) & Toilet paper roll wrapped in self-adhesive wrap  & Move the roll between the center target and each target on the shelf, resting between each motion and at the end \\ \hline
\end{tabular}}

\label{tab:act_descp_2}
\end{table*}

% \begin{table}[h]
% \begin{center}
% \begin{minipage}{299pt}
% \caption{StrokeRehab activity description}\label{app_tab1_stroke_act}%
% \begin{tabular}{llllll}
% \toprule
% Activity   & Workspace & Target
% object(s) & Instructions & Note  \\
% \midrule
% brushing  & & & & &   \\
% combing   & & & & &  \\
% deodorant & & & & &   \\
% drinking & & & & &   \\
% face washing & & & & & \\
%face washing & Sink with a small tub (32.3 x 24.1 x 2.5 $cm^3$) in it and two folded washcloths on either side of the countertop, 30 cm from edge closest to patient  & Washcloths, faucet handle, and tub & Fill tub with water, dip washcloth on the right side into water, wring it, wiping each side of their face with wet washcloth, place it back on countertop. Use washcloth on the left side to dry face, place it back on countertop & \\
% feeding & & & & &  \\
% glasses & & & & &  \\
% shelf & & & & &   structured\\
% RTT & & & & &   structured\\
% \botrule
% \end{tabular}
% \footnotetext{We show the held-out result for classifying KL grades(0,1,2,3,4) using different data-driven approaches.}
% \footnotetext[1]{Comparison against KL grade.} 

% \end{minipage}
% \end{center}
% \end{table} 

%\subsection{OAI-ZIB}
%{\color{red} TODO add additional data description}

\section{Additional Results}

\subsection{Quantification of Impairment in Stroke Patients} \label{sec_app:stroke_impairment}

\begin{figure}
\begin{tabular}{
 >{\centering\arraybackslash}m{0.3\linewidth} >{\centering\arraybackslash}m{0.3\linewidth}  >{\centering\arraybackslash}m{0.3\linewidth}   }
 Brushing & Combing & Deodorant \\
0.409 [0.158,0.610] & 0.599 [0.396,0.745] & 0.683 [0.510,0.803] \\
 \includegraphics[width=\linewidth]{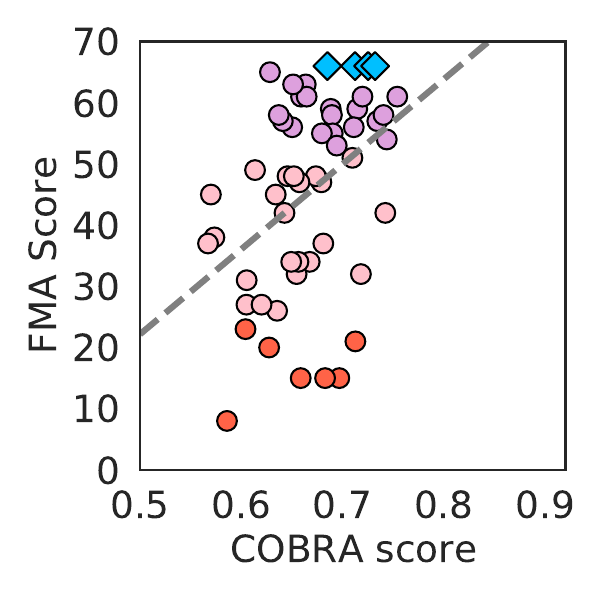} &\includegraphics[width=\linewidth]{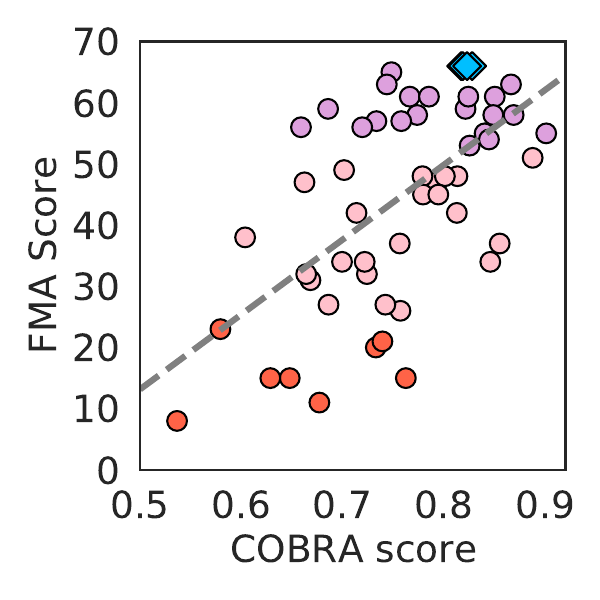} & \includegraphics[width=\linewidth]{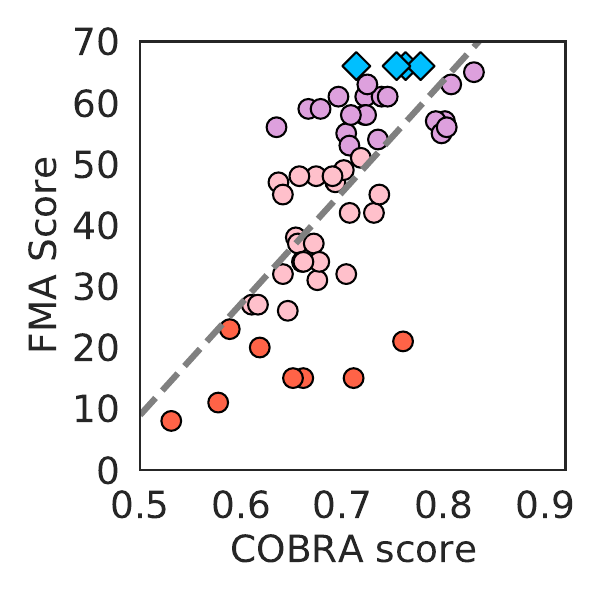} \\
  Drinking & Face-wash & Feeding \\
0.600 [0.398,0.746]& 0.397 [0.142,0.602] & 0.585 [0.376,0.737]\\ \includegraphics[width=\linewidth]{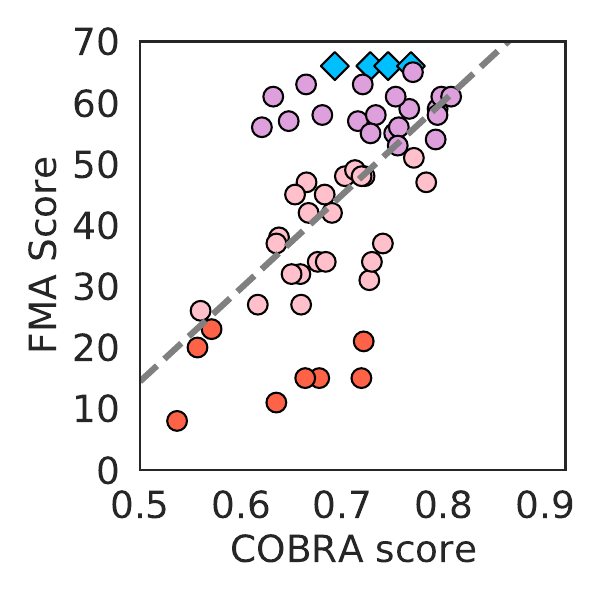} &   \includegraphics[width=\linewidth]{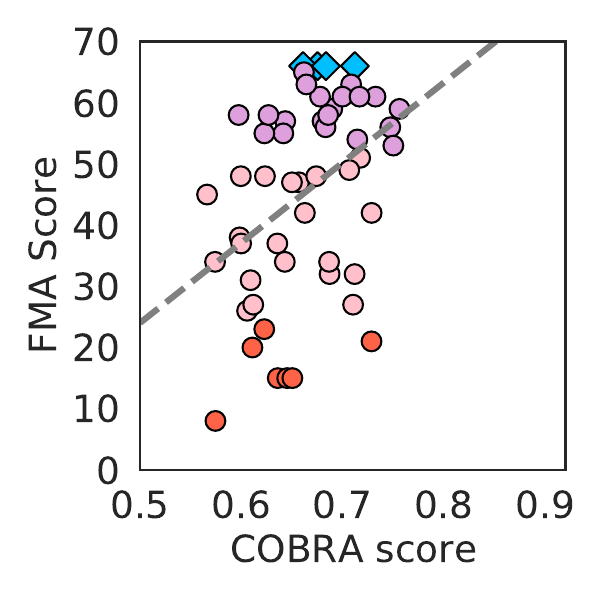} & \includegraphics[width=\linewidth]{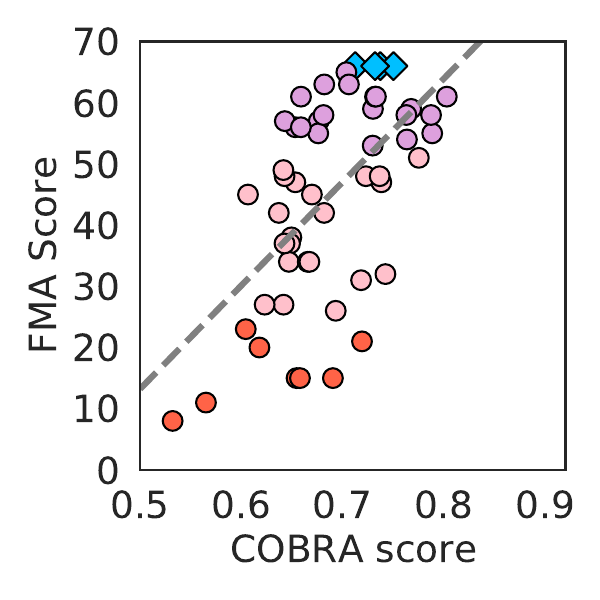} \\
 Glasses & Shelf & Table-top \\
0.745 [0.596,0.844] &  0.752 [0.605,0.850] & 0.849 [0.752,0.910]\\ \includegraphics[width=\linewidth]{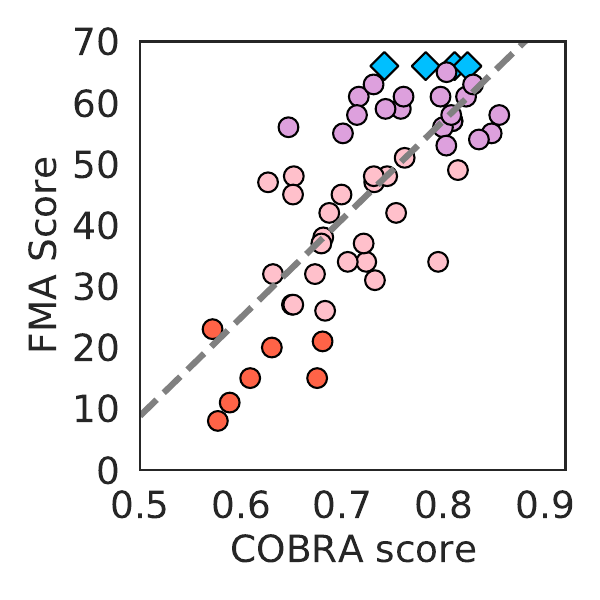} &   \includegraphics[width=\linewidth]{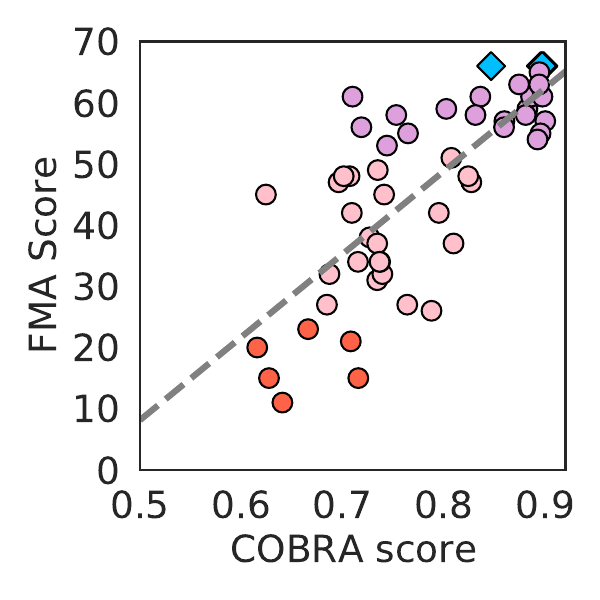} & \includegraphics[width=\linewidth]{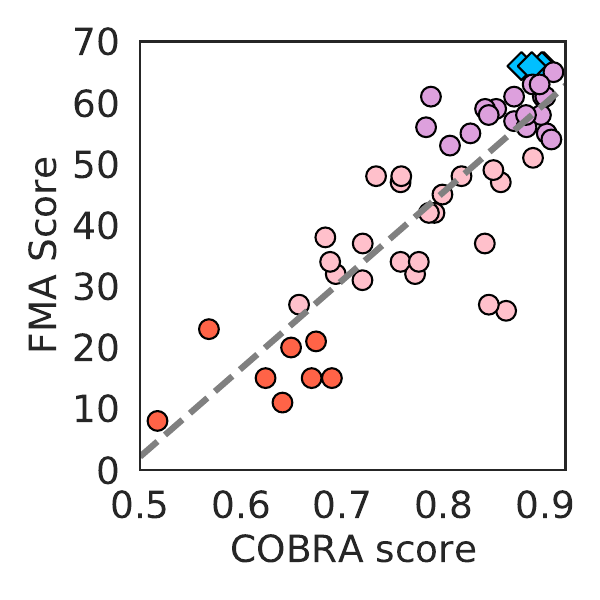} \\
\end{tabular}
\caption{\textbf{Correlation between wearable-sensor COBRA score and clinical assessment for individual rehabilitation activities.} Scatterplots of the Fugl-Meyer assessment (FMA) score, based on in-person examination by an expert, and the proposed data-driven COBRA score computed from wearable-sensor data for individual rehabilitation activities. The correlation coefficient $\rho$ is highest for simpler more structured activities such as Glasses, Shelf and Table-top.
}\label{fig:stroke_scatterplots_sensors}
\end{figure}

\begin{figure}
\begin{tabular}{
 >{\centering\arraybackslash}m{0.33\linewidth} >{\centering\arraybackslash}m{0.33\linewidth}  >{\centering\arraybackslash}m{0.33\linewidth}   }
 Brushing & Combing & Deodorant \\
0.521 [0.272,0.705] & 0.305 [0.016,0.547] & 0.542 [0.305,0.716] \\
\includegraphics[width=\linewidth]{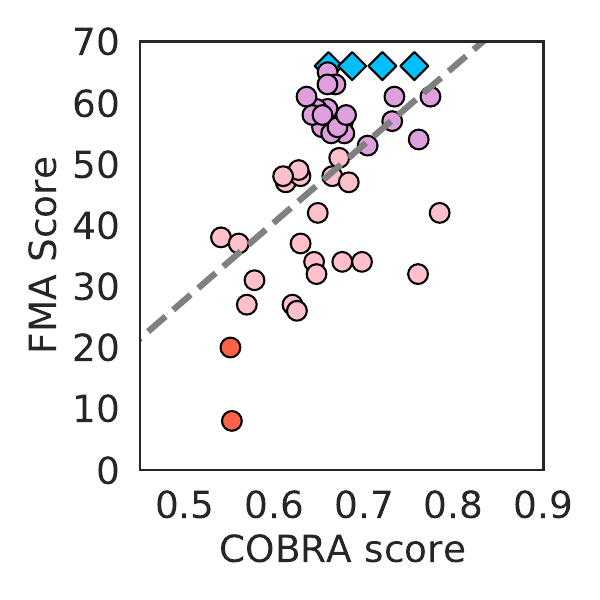} &\includegraphics[width=\linewidth]{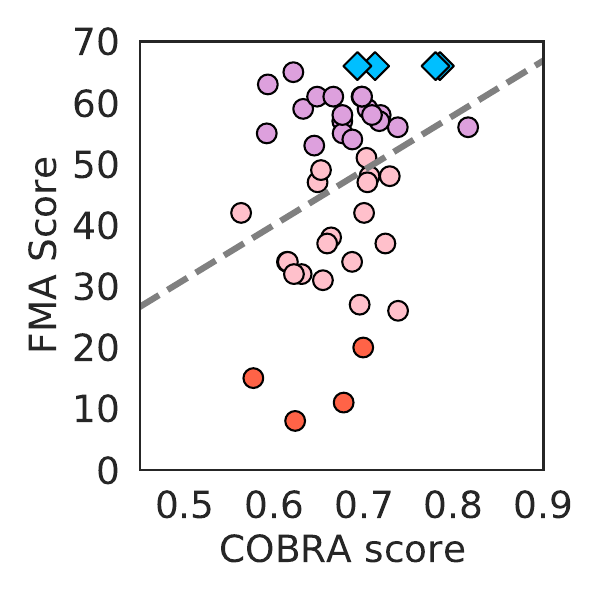} & \includegraphics[width=\linewidth]{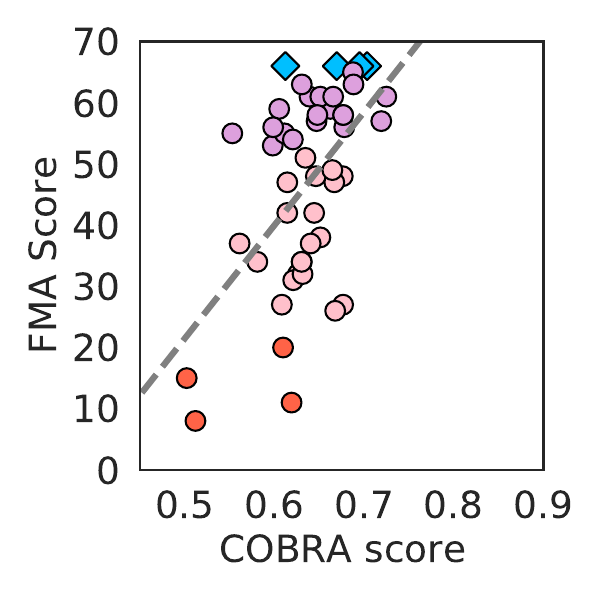} \\
Drinking & Face-wash & Feeding \\
0.500 [0.251,0.687] &  0.278 [-0.017,0.528] & 0.520 [0.280,0.699]\\ 
\includegraphics[width=\linewidth]{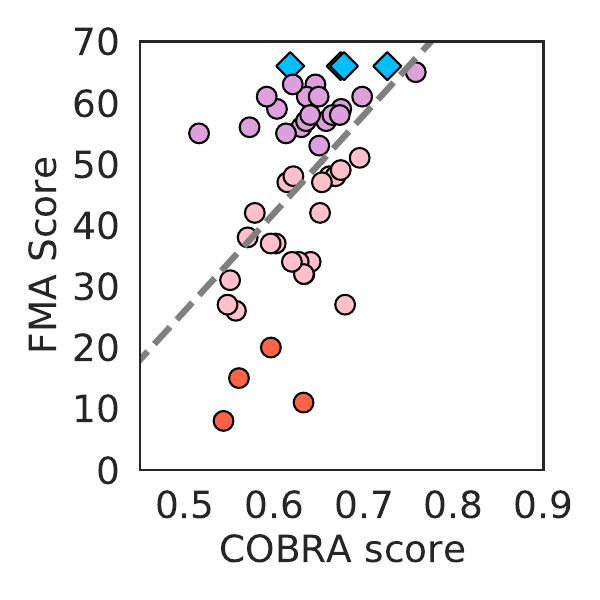} &   \includegraphics[width=\linewidth]{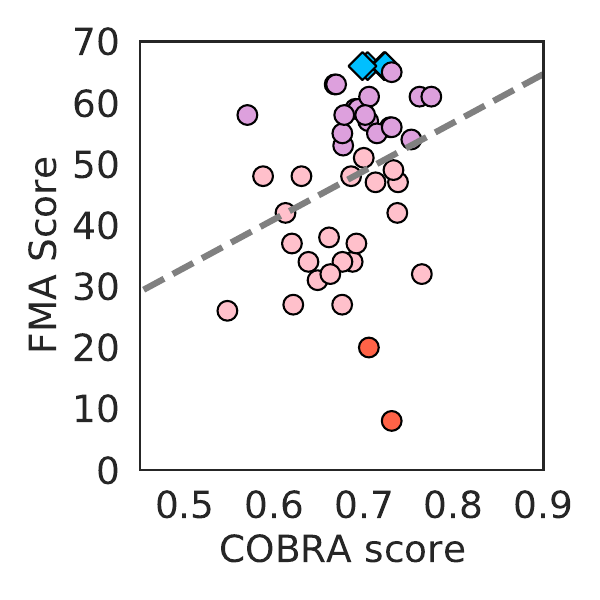} & \includegraphics[width=\linewidth]{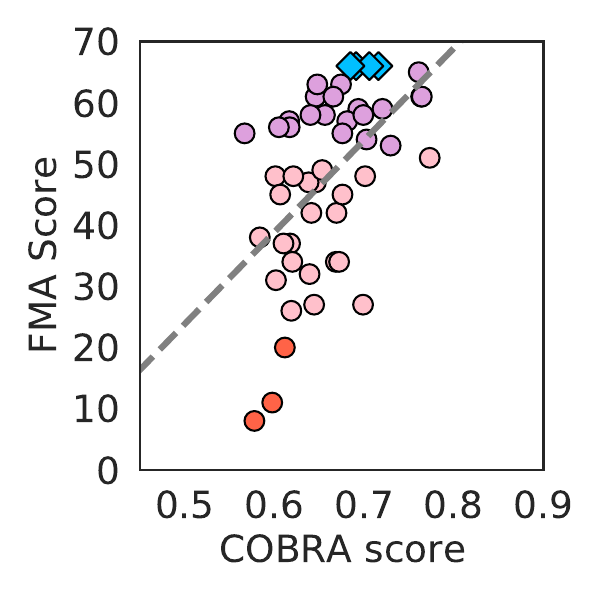} \\
Glasses & Shelf & Table-top \\
 0.650 [0.441,0.793] & 0.693 [0.509,0.816] & 0.615 [0.411,0.760]\\ \includegraphics[width=\linewidth]{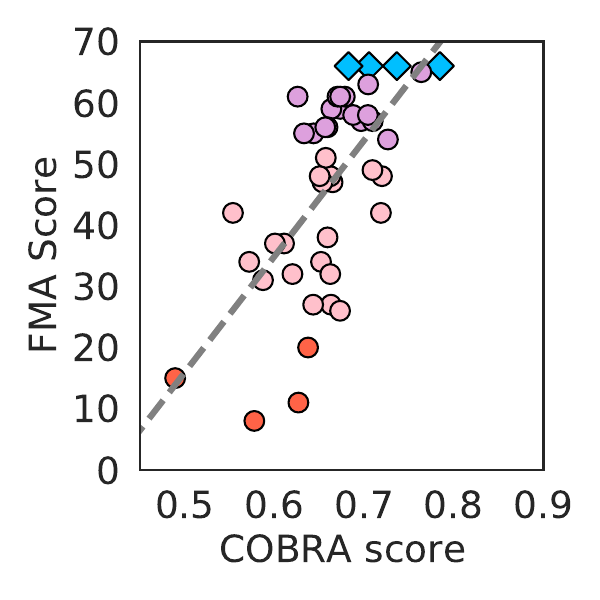} &   \includegraphics[width=\linewidth]{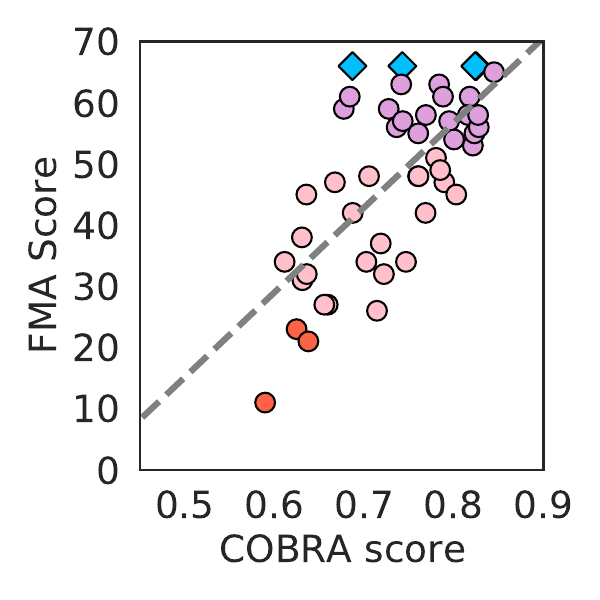} & \includegraphics[width=\linewidth]{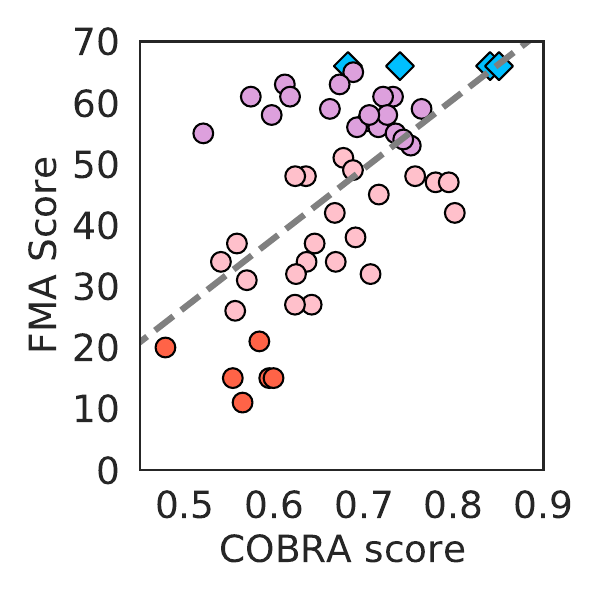} \\
\end{tabular}
\caption{
%COBRA scores on single activity for stroke impairment quantification computed from video data. Each individual figure compares data-driven COBRA score against the Fugl-Meyer assessment (FMA) score, based on in-person examination by an expert. Simple, more structured activities (Glasses, Shelf,Table-Top) have higher correlation coefficient $\rho$ than more complicated e wearable-sensor data.
\textbf{Correlation between video COBRA score and clinical assessment for individual rehabilitation activities.} Scatterplots of the Fugl-Meyer assessment (FMA) score, based on in-person examination by an expert, and the proposed data-driven COBRA score computed from video data for individual rehabilitation activities. The correlation coefficient $\rho$ is highest for simpler more structured activities such as Glasses, Shelf and Table-top.
}\label{fig:stroke_scatterplots_video}
\end{figure}

Figures~\ref{fig:stroke_scatterplots_sensors}, and \ref{fig:stroke_scatterplots_video} show scatterplots of the FMA and COBRA scores for each rehabilitation activity. Tables~\ref{app_tab_test_performance_sensor_healthy}, \ref{app_tab_test_performance_sensor_all_subjects}, and~\ref{app_tab_test_performance_video_healthy} report the accuracy and precision of the AI models described in Section~\ref{sec:COBRA_stroke}. 

%We provide the test performance of StrokeRehab clinically meaningful task on held-out healthy
%subjects in Table~\ref{app_tab_test_performance_sensor_healthy} (sensor) and Table~\ref{app_tab_test_performance_video_healthy} (video).
%In the main text, we discussed the result for characterizing impairment with all patient activities and the interesting Table-top activity. We now show COBRA scores derived from single activity and impairment quantification performance in Figure \ref{fig:stroke_scatterplots_sensors} (sensor) and Figure \ref{fig:stroke_scatterplots_video} (video). 

\begin{table}[h]
\caption{Performance of the AI models used to compute the COBRA score for quantification of stroke-induced impairment from wearable-sensor data on held-out healthy subjects. 95\% CIs are shown in brackets.}\label{app_tab_test_performance_sensor_healthy}
\centering
% \resizebox{1\linewidth}{!}{
\begin{tabular}{ccccc}
\toprule
Activity   & Motion-based & Motion-based  & Overall  & Overall \\
   &  Accuracy &  Precision &  Accuracy &  Precision \\
\midrule
All &\makecell{0.734\\ \lbrack0.726,0.747\rbrack} & \makecell{0.781\\\lbrack0.773,0.791\rbrack}& \makecell{0.755\\\lbrack0.747,0.761\rbrack} & \makecell{0.719\\\lbrack0.712,0.726\rbrack}    \\
\midrule
Brushing  &\makecell{0.588\\ \lbrack0.574,0.602\rbrack} & \makecell{0.613\\\lbrack0.598,0.625\rbrack}& \makecell{0.661\\\lbrack0.652,0.672\rbrack} & \makecell{0.616\\\lbrack0.605,0.624\rbrack}    \\
Combing  &\makecell{0.815\\\lbrack0.803,0.826\rbrack} & \makecell{0.804\\\lbrack0.794,0.816\rbrack}& \makecell{0.763\\\lbrack0.750,0.777\rbrack} & \makecell{0.735\\\lbrack0.725,0.746\rbrack}  \\
Deodorant & \makecell{0.692\\\lbrack0.681,0.702\rbrack}& \makecell{0.693\\\lbrack0.682,0.705\rbrack}& \makecell{0.718\\\lbrack0.708,0.726\rbrack}& \makecell{0.682\\\lbrack0.674,0.691\rbrack}   \\
Drinking & \makecell{0.696\\ \lbrack0.686,0.708\rbrack}& \makecell{0.740\\ \lbrack0.729,0.748\rbrack}& \makecell{0.743\\ \lbrack0.736,0.751\rbrack} & \makecell{0.710\\ \lbrack0.703,0.716\rbrack}  \\
Face-wash & \makecell{0.605\\\lbrack0.591,0.616\rbrack} & \makecell{0.628\\\lbrack0.618,0.638\rbrack}& \makecell{0.585\\\lbrack0.572,0.593\rbrack}& \makecell{0.569\\\lbrack0.561,0.576\rbrack}\\
Feeding & \makecell{0.623\\ \lbrack0.608,0.642\rbrack} & \makecell{0.643\\ \lbrack0.625,0.662\rbrack}& \makecell{0.676\\ \lbrack0.662,0.688\rbrack}&  \makecell{0.653\\ \lbrack0.642,0.665\rbrack}\\
Glasses &\makecell{0.781\\\lbrack0.773,0.790\rbrack} & \makecell{0.768\\\lbrack0.759,0.775\rbrack}& \makecell{0.699\\ \lbrack0.690,0.709\rbrack}& \makecell{0.686\\\lbrack0.678,0.693\rbrack} \\
Shelf & \makecell{0.831\\ \lbrack0.822,0.838\rbrack}& \makecell{0.927\\ \lbrack0.922,0.932\rbrack}& \makecell{0.835\\ \lbrack0.827,0.843\rbrack}& \makecell{0.773\\ \lbrack0.765,0.779\rbrack}\\
Table-top & \makecell{0.807\\\lbrack 0.801,0.813\rbrack}& \makecell{0.874\\ \lbrack0.868,0.879\rbrack}& \makecell{0.746\\ \lbrack0.736,0.759\rbrack}& \makecell{0.716\\ \lbrack0.710,0.723\rbrack}\\
\botrule
\end{tabular}
% }
% \footnotetext{We show the held-out result on for classifying primitives using wearable-sensor data. 95\% CI via bootstrap is shown in the brackets.}

% \end{minipage}
% \end{center}
\end{table}

\begin{table}[h]
\caption{Performance of the AI models used to compute the COBRA score for quantification of stroke-induced impairment from wearable-sensor data on  subjects with different levels of impairment. Performance degrades as the impairment level increases. 95\% CIs are shown in brackets.}\label{app_tab_test_performance_sensor_all_subjects}
\centering
% \resizebox{1\linewidth}{!}{
\begin{tabular}{ccccc}
\toprule
Impairment   & Motion-based & Motion-based  & Overall  & Overall \\
 Level  &  Accuracy &  Precision &  Accuracy &  Precision \\
\midrule
Healthy &\makecell{0.747\\ \lbrack0.738,0.757\rbrack} & \makecell{0.773\\\lbrack0.763,0.783\rbrack}& \makecell{0.753\\\lbrack0.743,0.760\rbrack} & \makecell{0.725\\\lbrack0.716,0.732\rbrack}    \\
\midrule
Mild  &\makecell{0.707\\ \lbrack0.696,0.716\rbrack} & \makecell{0.689\\\lbrack0.678,0.697\rbrack}& \makecell{0.674\\\lbrack0.666,0.681\rbrack} & \makecell{0.669\\\lbrack0.661,0.676\rbrack}    \\
Moderate  &\makecell{0.541\\\lbrack0.532,0.551\rbrack} & \makecell{0.536\\\lbrack0.535,0.551\rbrack}& \makecell{0.545\\\lbrack0.539,0.552\rbrack} & \makecell{0.570\\\lbrack0.562,0.578\rbrack}  \\
Severe & \makecell{0.354\\\lbrack0.342,0.367\rbrack}& \makecell{0.333\\\lbrack0.321,0.346\rbrack}& \makecell{0.373\\\lbrack0.364,0.383\rbrack}& \makecell{0.395\\\lbrack0.386,0.405\rbrack}   \\
\botrule
\end{tabular}
\end{table}

\begin{table}[h]
\begin{center}
\begin{minipage}{323pt}
\caption{
% \textbf{Held-out functional motion prediction performance on healthy subjects (video)}. We show the model is able to capture healthy movement patterns. We display 95\% CI via bootstrap is shown in the brackets.}\label{app_tab_test_performance_video_healthy}%
Performance of the AI models used to compute the COBRA score for quantification of stroke-induced impairment from video data on held-out healthy subjects. 95\% CIs are shown in brackets.}\label{app_tab_test_performance_video_healthy}
\centering
% \resizebox{1\linewidth}{!}{
\begin{tabular}{ccccc}
\toprule
Activity   & Motion-based & Motion-based  & Overall  & Overall \\
   &  Accuracy &  Precision &  Accuracy &  Precision \\
\midrule
% \centering
% \resizebox{1\linewidth}{!}{
% \begin{tabular}{lllll}
% \toprule
% Activity   & Motion-based Acc & Motion-based Precision & Overall Acc & Overall Precision\\
% \midrule
All  &\makecell{0.732\\ \lbrack0.720,0.740\rbrack} & \makecell{0.654\\ \lbrack0.643,0.663\rbrack}& \makecell{0.608\\ \lbrack0.597,0.616\rbrack} & \makecell{0.661\\ \lbrack0.650,0.672\rbrack}     \\
\midrule
Brushing  &\makecell{0.781\\ \lbrack0.766,0.795\rbrack} & \makecell{0.709\\ \lbrack0.692,0.720\rbrack}& \makecell{0.611\\ \lbrack0.599,0.622\rbrack} & \makecell{0.670\\ \lbrack0.656,0.680\rbrack}     \\
Combing  &\makecell{0.714\\ \lbrack0.706,0.724\rbrack} & \makecell{0.817\\ \lbrack0.803,0.831\rbrack}& \makecell{0.547\\ \lbrack0.540,0.556\rbrack} & \makecell{0.631\\ \lbrack0.623,0.639\rbrack}     \\
Deodorant  &\makecell{0.629\\ \lbrack0.618,0.643\rbrack} & \makecell{0.512\\ \lbrack0.497,0.529\rbrack} & \makecell{0.443\\ \lbrack0.435,0.453\rbrack} & \makecell{0.576\\ \lbrack0.562,0.590\rbrack}     \\
Drinking  &\makecell{0.619\\ \lbrack0.605,0.633\rbrack} & \makecell{0.542\\ \lbrack0.527,0.557\rbrack}& \makecell{0.493\\ \lbrack0.482,0.502\rbrack} & \makecell{0.539\\ \lbrack0.526,0.551\rbrack}     \\
Face-wash  &\makecell{0.648\\ \lbrack0.635,0.661\rbrack} & \makecell{0.645\\ \lbrack0.633,0.660\rbrack}& \makecell{0.469\\ \lbrack0.461,0.478\rbrack} & \makecell{0.560\\ \lbrack0.551,0.573\rbrack}     \\
Feeding  &\makecell{0.524\\ \lbrack0.502,0.541\rbrack} & \makecell{0.421\\ \lbrack0.399,0.446\rbrack}& \makecell{0.467\\ \lbrack0.454,0.478\rbrack} & \makecell{0.494\\ \lbrack0.477,0.508\rbrack}     \\
Glasses  &\makecell{0.708\\ \lbrack0.699,0.715\rbrack} & \makecell{0.673\\ \lbrack0.399,0.446\rbrack}& \makecell{0.513\\ \lbrack0.454,0.478\rbrack} & \makecell{0.667\\ \lbrack0.477,0.508\rbrack}     \\
Shelf  &\makecell{0.717\\ \lbrack0.709,0.726\rbrack} & \makecell{0.633\\ \lbrack0.624,0.642\rbrack}& \makecell{0.556\\ \lbrack0.506,0.519\rbrack} & \makecell{0.612\\ \lbrack0.568,0.655\rbrack}     \\
Table-top  &\makecell{0.768\\ \lbrack0.760,0.777\rbrack} & \makecell{0.693\\ \lbrack0.683,0.702\rbrack}& \makecell{0.614\\ \lbrack0.605,0.623\rbrack} & \makecell{0.620\\ \lbrack0.603,0.637\rbrack}     \\
\botrule
\end{tabular}
% \footnotetext{We show the held-out result for classifying primitives using video data. 95\% CI via bootstrap is shown in the brackets.}
\end{minipage}
\end{center}
\end{table}

\begin{table}[h]
\caption{
% \textbf{Held-out functional motion prediction performance on all subjects (video)}. We show the model is able to capture healthy movement patterns. For stroke patients, we see a decreasing performance. We display 95\% CI via bootstrap is shown in the brackets.}\label{app_tab_test_performance_video_all_subjects}
% \centering
% % \resizebox{1\linewidth}{!}{
% \begin{tabular}{lllll}
% \toprule
% Group   & Motion-based Acc & Motion-based Precision & Overall Acc & Overall Precision\\
% \midrule
Performance of the AI models used to compute the COBRA score for quantification of stroke-induced impairment from video data on subjects with different levels of impairment. Performance degrades as the impairment level increases. 95\% CIs are shown in brackets.}\label{app_tab_test_performance_video_all_subjects}
\centering
% \resizebox{1\linewidth}{!}{
\begin{tabular}{ccccc}
\toprule
Impairment   & Motion-based & Motion-based  & Overall  & Overall \\
 Level  &  Accuracy &  Precision &  Accuracy &  Precision \\
\midrule
Healthy &\makecell{0.732\\ \lbrack0.722,0.742\rbrack} & \makecell{0.654\\\lbrack0.643,0.664\rbrack}& \makecell{0.608\\\lbrack0.599,0.618\rbrack} & \makecell{0.662\\\lbrack0.652,0.672\rbrack}    \\
\midrule
Mild  &\makecell{0.615\\ \lbrack0.605,0.627\rbrack} & \makecell{0.501\\\lbrack0.492,0.512\rbrack}& \makecell{0.505\\\lbrack0.497,0.513\rbrack} & \makecell{0.551\\\lbrack0.542,0.561\rbrack}    \\
Moderate  &\makecell{0.555\\\lbrack0.546,0.566\rbrack} & \makecell{0.421\\\lbrack0.411,0.432\rbrack}& \makecell{0.441\\\lbrack0.433,0.448\rbrack} & \makecell{0.496\\\lbrack0.487,0.505\rbrack}  \\
Severe & \makecell{0.425\\\lbrack0.410,0.439\rbrack}& \makecell{0.348\\\lbrack0.338,0.356\rbrack}& \makecell{0.352\\\lbrack0.341,0.362\rbrack}& \makecell{0.339\\\lbrack0.330,0.348\rbrack}   \\
\botrule
\end{tabular}
\end{table}

% \begin{figure}[h]%
% \centering
% \includegraphics[width=0.9\textwidth]{figures/appdx - corr per act.pdf}
% \caption{\textbf{Using single activity to develop COBRA scores further improves data efficiency.} Correlation between COBRA scores and FMA is shown black squares in with error bars indicating 95\% confidence intervals. Number of sensor data sequences per subject is shown in orange bars. Structured activities(RTT and shelf) show good quantification results with far fewer input data.}\label{app_corr_per_act}
% \end{figure}

% \begin{figure}[h]%
% \centering
% \includegraphics[width=0.9\textwidth]{figures/appdx - sensor_train_combined_eval_single.pdf}
% \caption{COBRA scores on single activity for stroke impairment quantification using sensor data.}\label{fig:stroke_scatterplots_sensors}
% \end{figure}

% \begin{figure}[h]%
% \centering
% \includegraphics[width=0.9\textwidth]{figures/appdx - video_train_combined_eval_single.pdf}
% \caption{COBRA scores on single activity for stroke impairment quantification using video data.}\label{fig:stroke_scatterplots_video}
% \end{figure}

\begin{figure}
%\hspace{-0.5cm}
\begin{tabular}{
 >{\centering\arraybackslash}m{0.3\linewidth} >{\centering\arraybackslash}m{0.3\linewidth}  >{\centering\arraybackslash}m{0.3\linewidth}   }
  & Blurred & Unblurred  \\
 Unblurred + Blurred &\includegraphics[width=0.5\linewidth]{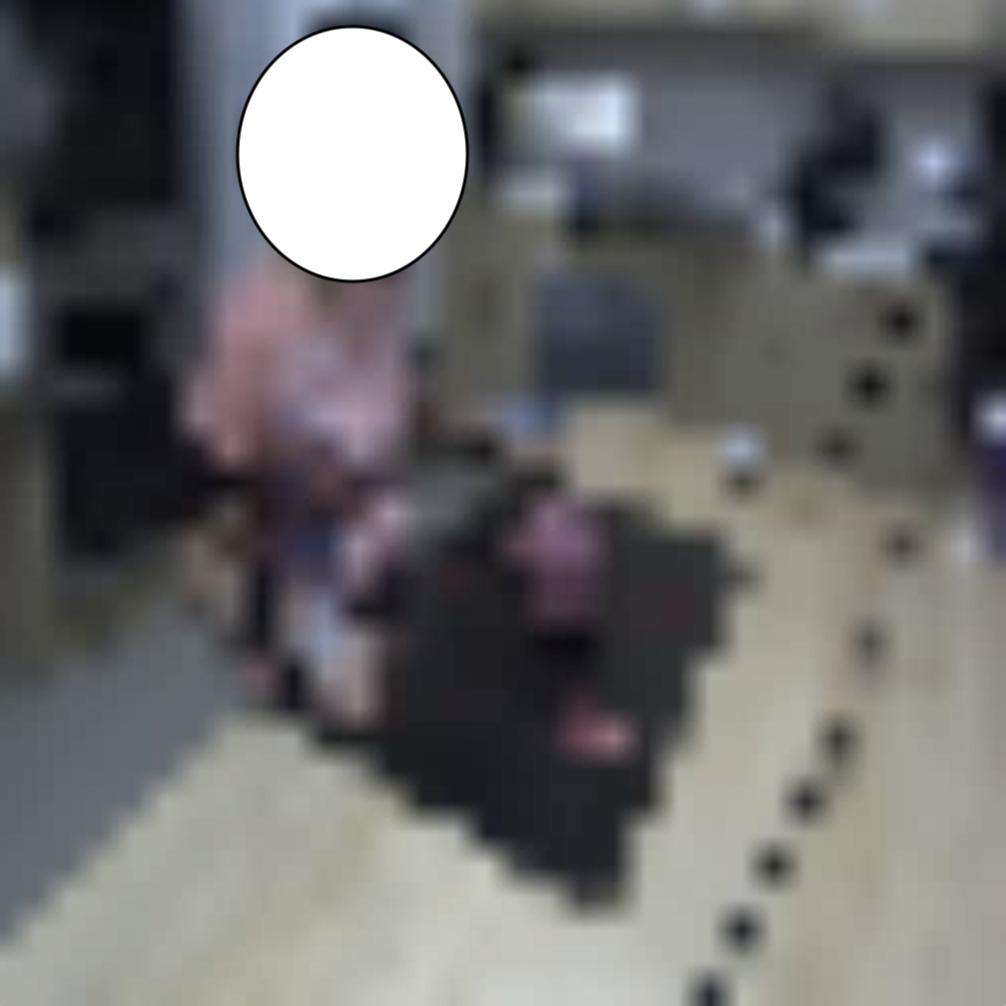} & \includegraphics[width=0.5\linewidth]{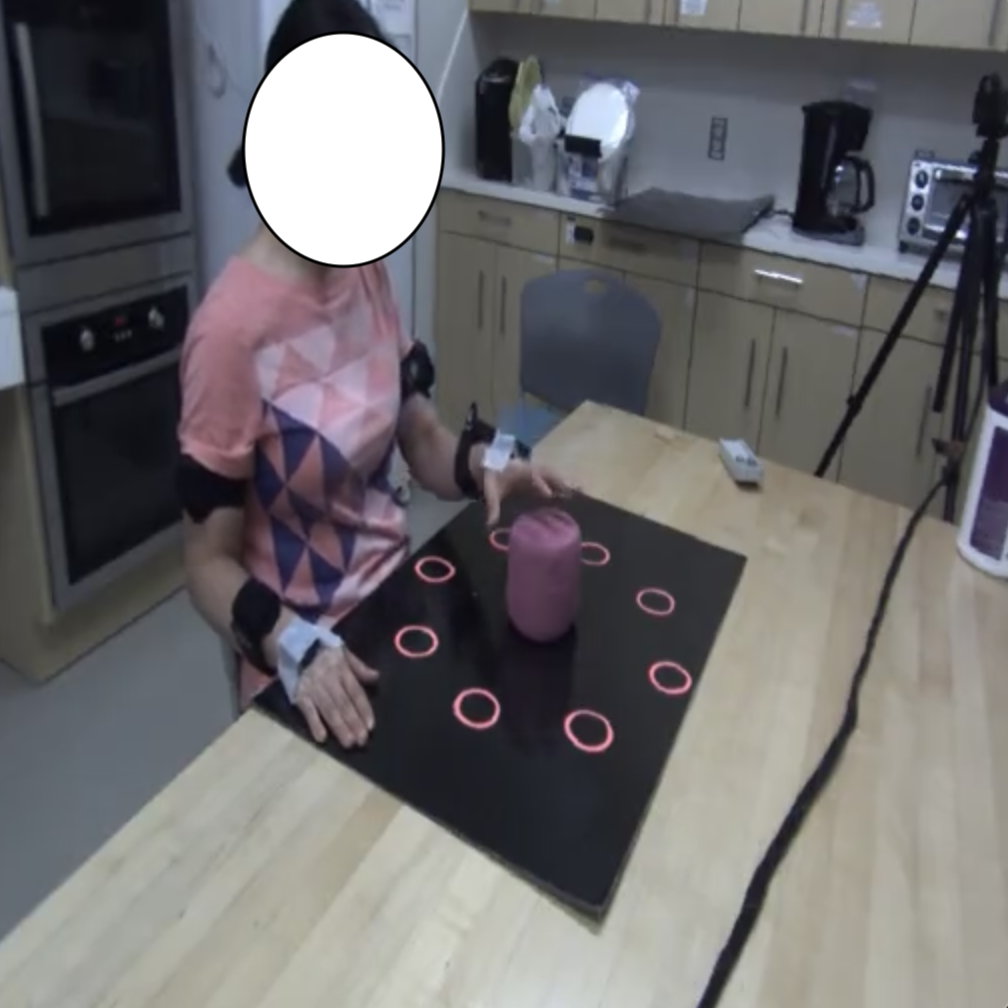} \\
  $\rho$ = 0.582 &  $\rho$ = 0.584 & $\rho$ = 0.817  \\
 {\small 95$\%$ CI [0.374,0.734]} 
 & {\small 95$\%$ CI [0.270,0.786]} 
 & {\small 95$\%$ CI [0.634,0.913]} \\
 \includegraphics[width=\linewidth]{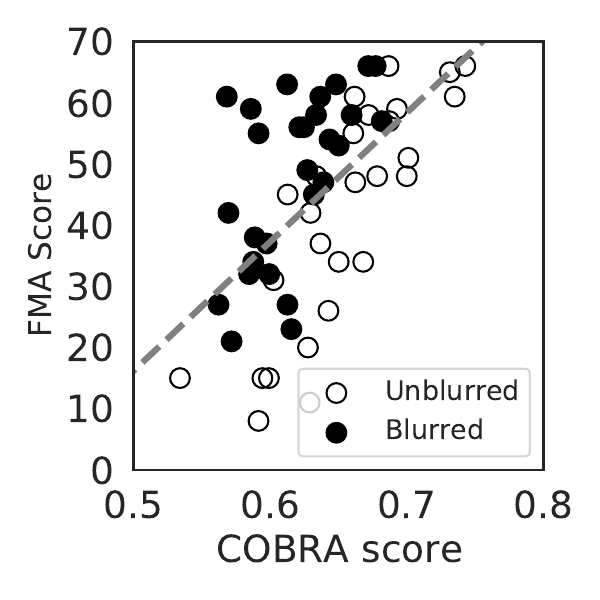} &   \includegraphics[width=\linewidth]{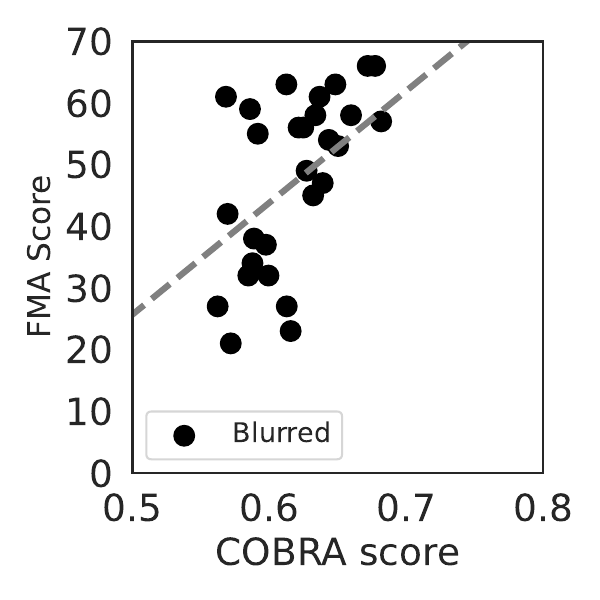} & \includegraphics[width=\linewidth]{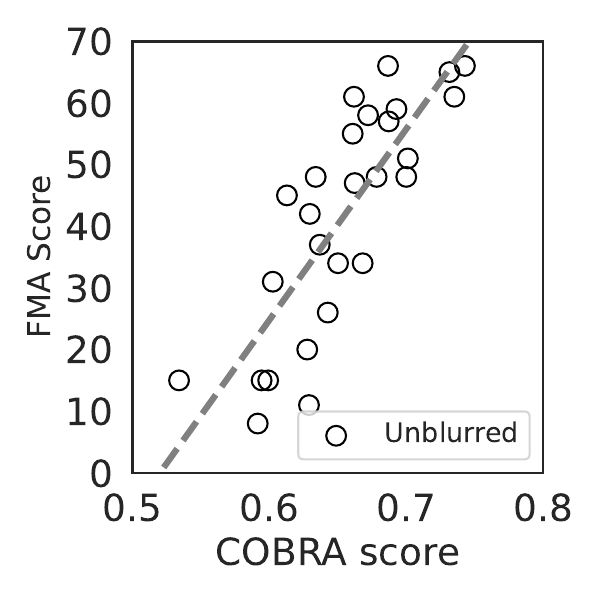} \\
\end{tabular}
\caption{\textbf{Video quality as a confounding factor for the video-based COBRA score.} Vision models are sensitive to video quality. One example is the blurring effect. The bottom left scatterplot shows the COBRA score computed from blurred and non-blurred videos, as well as the corresponding Fugl-Meyer assessment (FMA) score. Blurring decreases model confidence, systematically decreasing the COBRA score. The bottom middle and right scatterplots show that stratifying the videos according to their quality corrects for the confounding factor, improving the correlation coefficient $\rho$ between the COBRA and FMA scores.}\label{app_fig_blur}
\end{figure}

In Figure~\ref{fig:light_dark} we show that object color is a confounding factor, which can spuriously reduce model confidence and therefore distort the COBRA score. To complement this observation, we analyzed the impact of varying video resolution on the COBRA score. We blurred half of the videos (chosen at random), reducing their resolution by a factor of 16 along each axis and then restoring them to their original dimensions. Figure~\ref{app_fig_blur} shows the results of applying the COBRA score to a dataset containing the blurred and non-blurred videos. Blurring acts as a confounding factor, producing a spurious decrease in model confidence independent of impairment, which reduces the correlation between FMA and the COBRA score. This can be corrected by stratifying the videos, separating them according to whether they are blurred or not.
%In Figure~\ref{fig:light_dark} we show that object color is main text, we demonstrate that confounding factors can artificially reduce model confidence. We present an example in Table-top activity Figure ~ \ref{fig:light_dark}, where we examine how visual factors, specifically object color, affect the COBRA score derived from video data. Additionally, we analyze the impact of video corruption on model confidence. To illustrate this, we blur all videos of stroke subjects using a down-scale resizing approach, reducing the frame size by 16 times along each axis and then restoring it to the original shape. Our results reveal that even simple video corruption can lead to a spurious decrease in model confidence Figure~\ref{app_fig_blur}.

% \begin{figure}[h]%
% \centering
% \includegraphics[width=0.9\textwidth]{figures/appdx - blur.pdf}
% \caption{\textbf{Be careful with confounding factors.} 
% the video quality is another confounding factor that could affect stroke impairment quantification as model gives low confidence to corrupted(blurry) videos.}\label{app_fig_blur}
% \end{figure} 

\subsection{Quantification of Knee-Osteoarthritis Severity} \label{sec_app:Knee_OA}

%The test performance of Knee OA clinically meaningful task on held-out healthy
%subjects are reported in 
Table \ref{app_tab_test_performance_knee} reports the pixel-wise accuracy and precision of the AI model described in Section~\ref{sec:COBRA_knee}.
%used to compute the COBRA score for quantification of knee-osteoarthritis severity from MRI scans on held-out healthy subjects..

% \begin{table}[h]
% \begin{center}
% \begin{minipage}{323pt}
% \caption{Knee OA clinically meaningful task held-out test performance on healthy subjects. We show the held-out result for segmenting tissue types. 95\% CI via bootstrap is shown in the brackets.}\label{app_tab_test_performance_knee}%
% \centering
% \resizebox{0.45\linewidth}{!}{
% \begin{tabular}{lllll}
% \toprule
% Tissue   & Dice coefficient\\
% \midrule
% All &\makecell{0.732\\ \lbrack0.720,0.740\rbrack}  \\
% \midrule
% Bone  &\makecell{0.732\\\lbrack0.720,0.740\rbrack}  \\
% Cartilage  &\makecell{0.732\\\lbrack0.720,0.740\rbrack}  \\
% \midrule
% Femur bone  &\makecell{0.732\\\lbrack0.720,0.740\rbrack}  \\
% Femur cartilage  &\makecell{0.781\\\lbrack0.766,0.795\rbrack}\\
% Tibia bone  &\makecell{0.714\\\lbrack0.706,0.724\rbrack} \\
% Tibia cartilage  &\makecell{0.629\\\lbrack0.618,0.643\rbrack} \\
% \botrule
% \end{tabular}
% }
% % \footnotetext{We show the held-out result for segmenting tissue types. 95\% CI via bootstrap is shown in the brackets.}
% \end{minipage}
% \end{center}
% \end{table}

\begin{table}[h]
\caption{Pixel-wise performance of the AI models used to compute the COBRA score for quantification of knee-osteoarthritis severity from MRI scans on held-out healthy subjects. 95\% CIs are shown in brackets.
%Knee OA clinically meaningful task held-out test performance. We show the held-out result for pixel-wise classification metrics(overall accuracy and precision excluding background class) within each KL group. 95\% CI via Fisher transformation is shown in the brackets.
}\label{app_tab_test_performance_knee}%
\centering
\begin{tabular}{lll}
\toprule
Group   & Accuracy & Precision\\
\midrule
KL = 0 ( Healthy )  &\makecell{0.995\\ \lbrack0.993,0.997\rbrack} & \makecell{0.922\\ \lbrack0.878,0.952\rbrack}  \\
\midrule
KL = 1 ( Doubtful )  &\makecell{0.994\\\lbrack0.992,0.996\rbrack} &\makecell{0.922\\ \lbrack0.890,0.956\rbrack}  \\
\midrule
KL = 2 ( Minimal )  &\makecell{0.994\\\lbrack0.990,0.996\rbrack} &\makecell{0.907\\ \lbrack0.852,0.945\rbrack}  \\
\midrule
KL = 3 ( Moderate ) &\makecell{0.992\\\lbrack0.983,0.996\rbrack} &\makecell{0.885\\ \lbrack0.816,0.931\rbrack}  \\
\midrule
KL = 4 ( Severe ) &\makecell{0.991\\\lbrack0.979,0.995\rbrack} & \makecell{0.851\\ \lbrack0.721,0.917\rbrack}  \\
\botrule
\end{tabular}
% \footnotetext{We show the held-out result for segmenting tissue types. 95\% CI via bootstrap is shown in the brackets.}
\end{table}

%To evaluate discriminative performance in classifying KL grades, we compare a logistic regression model using COBRA score in different tissues with traditional deep learning approach. Our method show advantage using anomaly detection.

% \begin{table}[h]
% \begin{center}
% \begin{minipage}{293pt}
% \caption{KL grade classification is a scenario when our methods show efficient quantification results}\label{app_tab2}%
% \begin{tabular}{lll}
% \toprule
% (Knee OA KL discrimination)\footnotemark[1]   & Balanced Accuracy & F1 score  \\
% \midrule
% 2D CNN   & 0.21  & 0.20 \\
% 3D CNN   & 0.20  & 0.19 \\
% \midrule
% LR using COBRA confidence(all tissues)   & 0.53  & 0.47 \\
% LR using COBRA confidence(cartilage)  & 0.44  & 0.38 \\
% \botrule
% \end{tabular}
% \footnotetext{We show the held-out result for classifying KL grades(0,1,2,3,4) using different data-driven approaches.}
% \footnotetext[1]{Comparison against KL grade.} 

% \end{minipage}
% \end{center}
% \end{table} 
\clearpage

\section{Distance-based Anomaly Quantification}\label{sec:distance_based}

%\subsection{Method}
In this section we present an alternative method for anomaly detection and quantification that utilizes an AI model trained only on healthy patients. The method is based on the Fréchet Inception Distance (FID)~\cite{heusel2017gans} to quantify the deviation between a subject and a healthy population. FID is a metric designed to evaluate the similarity between two sets of feature representations extracted by a deep neural network. It has been applied to image generation~\cite{heusel2017gans}, where the goal is to determine whether generated images are close to real images or not. 

We propose to leverage FID to compare a potentially impaired subject to a healthy reference population using the same model features as in the COBRA framework. First, the data associated with all individuals is fed into a deep neural network, trained to perform a task relevant to the impairment or disease of interest. Then, the features extracted by the neural network are compared via FID to determine to what extent the subject deviates from the population.

%features of a healthy subject population with those extracted from an impaired subject, such as patients with a particular disease or condition. The idea is to assess whether the features of the impaired subject are significantly different from those of the healthy population, and if so, to what extent.

Let $x_1, x_2, ..., x_n$ be the features associated with the healthy reference subjects, and $y_1, y_2, ..., y_m$ the feature representations of the potentially-impaired subject. The sample mean and covariance matrices of these features are denoted by $\mu_x, \mu_y$ and $\Sigma_x, \Sigma_y$, respectively. The FID between the healthy population and the impaired subject is 
\begin{align}
    \operatorname{FID}(x, y) = \|\mu_x - \mu_y\|_2^2 + \operatorname{Trace}(\Sigma_x + \Sigma_y - 2(\Sigma_x \Sigma_y)^{1/2}),
\end{align} 
where $\operatorname{Trace}(\cdot)$ denotes the trace operator and $\|\cdot\|_2$ is the $\ell_2$ norm. The lower the FID between the two sets of features, the more similar they are.

%To apply the FID distance to our problem, we, first, extract the feature representations of the images/sensor data from the two populations using a pre-trained deep neural network. The pre-trained network is trained to identify the low-level semantic labels that are indicative of the impairment. For example, in the case of stroke-impairment quantification, the network is trained to predict the primitives performed by the subjects, whereas, in the case of estimating the degradation of cartilages, the network tries to segment different tissues and bones of the knee. 

As a proof of concept, we apply FID to quantification of stroke-induced impairment using the sensor dataset described in Section~\ref{sec:results_stroke}. 
% determine the impairment of individual subjects within the impaired population, we calculate the FID distance between their feature representations and the healthy population's feature representations. If a subject's feature representations are similar to those of the healthy population, their FID distance will be relatively low, suggesting that they are less impaired. Conversely, if a subject's feature representations are significantly different from those of the healthy population, their FID distance will be higher, suggesting that they are more impaired.
We apply the model described in Section~\ref{subsec:model_arch_wearable_sensors}, trained on the training cohort of healthy patients. The features used to compute the FID are extracted from the penultimate layer of the model. %, which dimensionality is 64 for each time step.
%To ensure an equitable comparison between healthy and stroke-impaired subjects, we employ a specific calibration method. 
Two held-out healthy subjects from the test cohort were randomly chosen to be the reference healthy population. The FID of the remaining subjects of the test cohort was computed with respect to this population.

%, namely 'C15' and 'C23,' in a random fashion, designating them as our baseline healthy population. Subsequently, all other subjects, including both stroke-impaired individuals and the remaining two healthy subjects ('C4' and 'C30'), are compared to this baseline representation. This approach enables us to discern how the distances between two healthy subjects compare to the distances between a stroke-impaired subject and a healthy subject, ensuring an accurate evaluation of impairments.

%\subsection{Results}
Table~\ref{tab:fid_distance_results} shows the correlation coefficient of the FID with respect to the reference healthy subjects and the Fugl-Meyer assessment (FMA) for different rehabilitation activities. The magnitude of the correlation is higher when motion primitives are utilized to compute the FID, and for more structured activities. The metrics are not as correlated as the COBRA score and FMA (see Figure~\ref{fig:stroke_scatterplots_sensors}), but these results suggest that there may be multiple ways of exploiting features extracted from AI models to perform anomaly detection and quantification.  
% Notably, all activities, particularly motion-based primitives, exhibit a high absolute correlation. This finding suggests that the FID distance-based approach can effectively quantify impairment. It demonstrates the potential of using the FID distance as a measure of impairment severity.

%This analysis focused exclusively on wearable-sensor data as a proof-of-concept. It demonstrates that our COBRA method is versatile enough to accommodate other approaches for evaluating feature representation distances to quantify impairment or severity.

\begin{table}[t]
\caption{\textbf{Correlation between Fréchet Inception Distance (FID) and clinical assessment.} Correlation coefficient between the FID (with respect to a reference healthy population) computed from wearable-sensor data and the Fugl-Meyer assessment for different rehabilitation activities. The metrics are more correlated when motion primitives are utilized to compute the FID, and for more structured activities. 95\% CIs are shown in brackets. %Spearman correlation also shows similar results
}\label{tab:fid_distance_results}%
\centering
% \resizebox{1\linewidth}{!}{
\begin{tabular}{llll}
\toprule
Activity   & Non-motion & Motion & All primitives\\
\midrule
All &\makecell{-0.359\\ \lbrack-0.461,-0.025\rbrack} & \makecell{-0.715\\\lbrack-0.780,-0.647\rbrack}& \makecell{-0.650\\\lbrack-0.719,-0.577\rbrack}    \\
\midrule
Brushing  &\makecell{-0.041\\ \lbrack-0.249,0.225\rbrack} & \makecell{-0.176\\\lbrack-0.426,0.001\rbrack}& \makecell{-0.139\\\lbrack-0.287,0.078\rbrack}   \\
Combing  &\makecell{-0.329\\\lbrack-0.434,-0.140\rbrack} & \makecell{-0.427\\\lbrack-0.703,-0.025\rbrack}& \makecell{-0.313\\\lbrack-0.469,-0.087\rbrack} \\
Deodorant & \makecell{-0.280\\\lbrack-0.500,-0.081\rbrack}& \makecell{-0.693\\\lbrack-0.793,-0.402\rbrack}& \makecell{-0.617\\\lbrack-0.712,-0.493\rbrack}   \\
Drinking & \makecell{-0.243\\ \lbrack-0.433,-0.032\rbrack}& \makecell{-0.352\\ \lbrack-0.477,-0.183\rbrack}& \makecell{-0.436\\ \lbrack-0.563,-0.277\rbrack}  \\
Face-wash & \makecell{-0.246\\\lbrack-0.433,-0.004\rbrack} & \makecell{-0.255\\\lbrack-0.464,0.025\rbrack}& \makecell{-0.347\\\lbrack-0.491,-0.217\rbrack}\\
Feeding & \makecell{-0.429\\ \lbrack-0.625,-0.230\rbrack} & \makecell{-0.396\\ \lbrack-0.561,-0.160\rbrack}& \makecell{-0.523\\ \lbrack-0.675,-0.329\rbrack}\\
Glasses &\makecell{-0.510\\\lbrack-0.619,-0.270\rbrack} & \makecell{-0.688\\\lbrack-0.760,-0.599\rbrack}& \makecell{-0.718\\ \lbrack-0.795,-0.624\rbrack} \\
Shelf & \makecell{-0.254\\ \lbrack-0.461,-0.034\rbrack}& \makecell{-0.684\\ \lbrack-0.760,-0.580\rbrack}& \makecell{-0.597\\ \lbrack-0.699,-0.439\rbrack}\\
Table-top & \makecell{-0.533\\\lbrack -0.640,-0.418\rbrack}& \makecell{-0.693\\ \lbrack-0.758,-0.605\rbrack}& \makecell{-0.609\\ \lbrack-0.694,-0.492\rbrack}\\
\botrule
\end{tabular}
% }
% \footnotetext{We show the held-out result on for classifying primitives using wearable sensor data. 95\% CI via bootstrap is shown in the brackets.}

% \end{minipage}
% \end{center}
\end{table}

\end{appendices}




\end{document}